\documentclass[10pt,twocolumn,letterpaper]{article}

\usepackage{cvpr}              %

\newcommand{\methodname}{ViCARD}
\newcommand{\methodfullname}{Visual Credit Assignment with Reward-Gated Distillation}

\newcommand{\samplingtemperature}{0.7}
\newcommand{\samplingtopp}{0.8}
\newcommand{\samplingtopk}{20}

\DeclareCaptionLabelFormat{tokenimportanceab}{#1~#2(a,b)}
\DeclareCaptionLabelFormat{tokenimportancec}{#1~#2(c)}
\DeclareCaptionLabelFormat{tokenimportancede}{#1~#2(d,e)}

\usepackage{placeins}
\usepackage{tabularx}
\usepackage{algorithm}
\usepackage{algpseudocode}
\algrenewcommand{\algorithmicrequire}{\textbf{Input:}}
\algrenewcommand{\algorithmicensure}{\textbf{Output:}}
\AtBeginDocument{\raggedbottom}
\makeatletter
\renewcommand\paragraph{\@startsection{paragraph}{4}{\z@}%
  {1pt plus 1pt minus 1pt}{-1em}{\normalfont\normalsize\bfseries}}
\makeatother

\usepackage{microtype}

\newcommand{\compactdisplayspacing}{%
  \setlength{\abovedisplayskip}{4pt plus 2pt minus 1pt}%
  \setlength{\belowdisplayskip}{4pt plus 2pt minus 1pt}}
\AddToHook{env/equation/begin}{\compactdisplayspacing}
\AddToHook{env/equation*/begin}{\compactdisplayspacing}

\definecolor{cvprblue}{rgb}{0.21,0.49,0.74}
\usepackage[pagebackref,breaklinks,colorlinks,allcolors=cvprblue]{hyperref}

\def\confName{CVPR}

\title{\methodname{}: \methodfullname{} for Multimodal Reasoning}

\author{Zijian Huang\\
University of Michigan\\
Ann Arbor, MI\\
{\tt\small zijianh@umich.edu}
\and
Jiasi Chen\\
University of Michigan\\
Ann Arbor, MI\\
{\tt\small jiasi@umich.edu}
}

\begin{document}
\maketitle
\flushbottom
\captionsetup[figure]{skip=4pt}
\setlength{\dbltextfloatsep}{3pt plus 2pt minus 1pt}
\begin{abstract}
Visual reasoning requires models to identify relevant visual evidence and use it to derive correct answers. Outcome-based reinforcement learning (RL) faces two key challenges in this regime: it yields no learning signal when perceptual errors cause all sampled responses to fail, and it broadcasts a uniform advantage across every token within a response. Conversely, while visual self-distillation offers dense guidance from evidence-centered image crops, student learning from token distributions alone often fails to elicit correct final answers. Motivated by an audit of training rollouts that exposes these complementary failure modes, we introduce \textbf{\methodname{}} (\underline{Vi}sual \underline{C}redit \underline{A}ssignment with \underline{R}eward-Gated \underline{D}istillation), which integrates per-token visual supervision with verified correctness rewards. For each rollout group, we leverage a teacher with its input image cropped to the relevant visual area, and use this to provide per-token supervision to the student while also valuing correctness. If all sampled rollouts fail, \methodname{} samples an additional crop-teacher answer and activates on-policy distillation only if that answer is correct. Together, these mechanisms provide fine-grained visual credit within informative reward groups and a learning signal for otherwise uninformative all-fail groups. Extensive comparisons across model scales demonstrate that \methodname{} improves fine-grained visual reasoning performance over RL and distillation baselines, while retaining broad visual reasoning capabilities.
\end{abstract}

\section{Introduction}
\label{sec:introduction}

\begin{figure*}[t]
\centering
\includegraphics[width=0.84\textwidth]{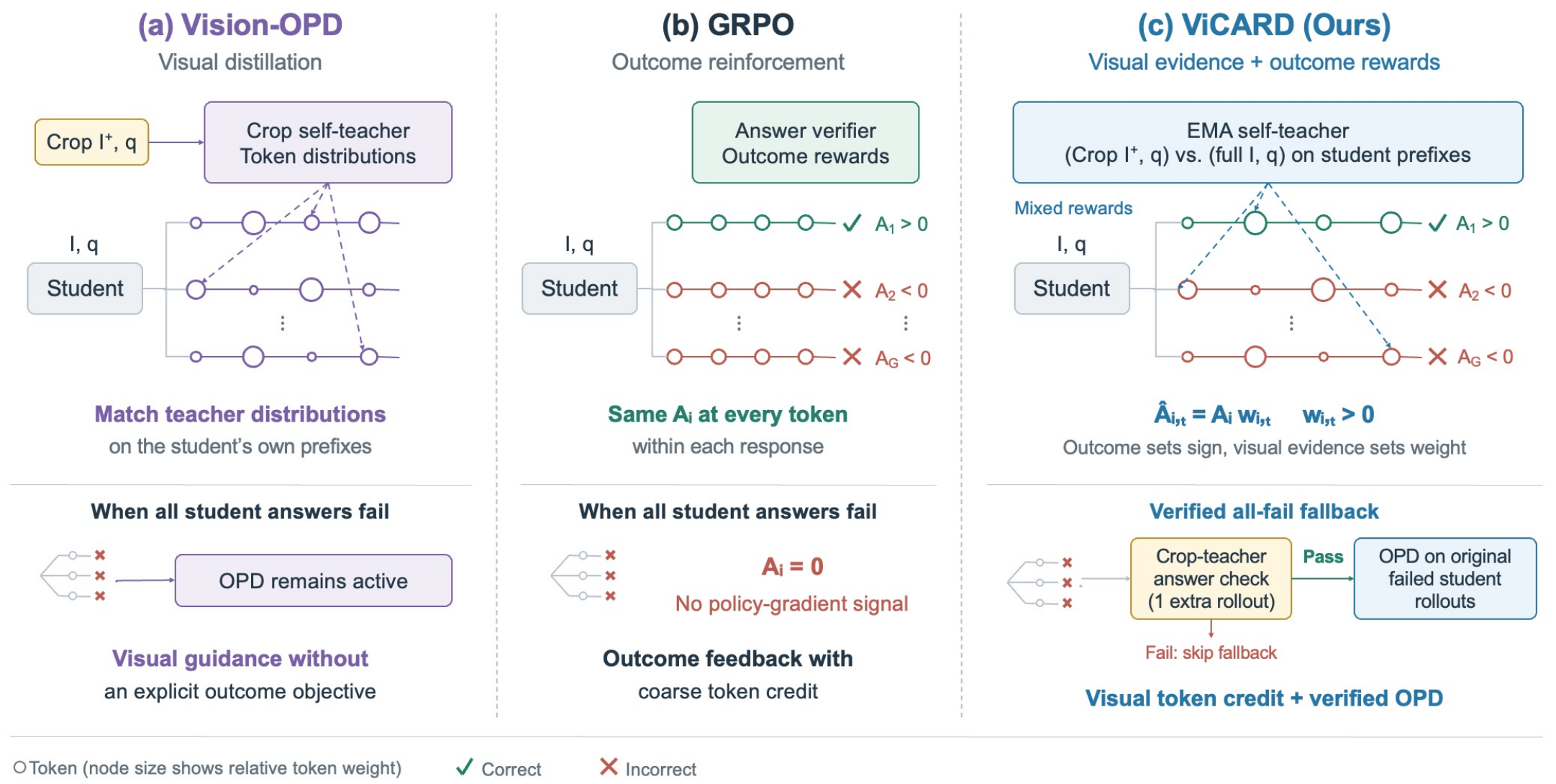}
\caption{\textbf{Comparison between Vision-OPD, GRPO, and \methodname{}.} All methods train on full-image student rollouts. Vision-OPD matches crop-teacher distributions in (a); GRPO assigns a shared outcome advantage to response tokens in (b). \methodname{} uses visual token weights in mixed-reward groups and verified OPD on original student rollouts when all answers fail in (c). Circle sizes indicate token weights; check marks and crosses denote correct and incorrect answers. Inference uses the full-image student alone.}
\label{fig:overall_comparison}
\end{figure*}

Reliable visual reasoning depends on both finding the right evidence and reasoning correctly from it. For instance, answering a query about a cluttered scene requires locating objects relevant to the query, just as reading a chart requires accurately mapping numerical values to their corresponding labels before reasoning from the numerical evidence. Even basic visual questions expose persistent perceptual failures in multimodal models~\cite{tong2024mmvp}. Recent advances in RL for vision-language models improve reasoning through verifiable rewards~\cite{wang2025vlrethinker,liu2025noisyrollout}, while visual perception methods help models recover small but decisive details~\cite{zheng2026deepeyes,wei2026zwz,yuan2026visionopd}. Learning to use those details reliably remains a central challenge for visual reasoning.

Group relative policy optimization (GRPO)~\cite{shao2024deepseekmath} has been adapted to visual reasoning~\cite{wang2025vlrethinker,liu2025noisyrollout}.
Vanilla GRPO samples several responses to the same question and reinforces those with higher relative rewards. However, when weak visual grounding causes every sampled rollout in a group to fail, the resulting rewards are uniformly zero and the outcome-driven policy gradient vanishes. Consequently, these \emph{all-fail groups} consume substantial rollout compute without contributing an outcome-based learning signal. In our audit of 1595 randomly selected questions from Vision-OPD's regional training dataset~\cite{yuan2026visionopd} using Qwen3.5-4B, all eight GRPO rollouts are incorrect on 22.13\% of questions, which is a significant minority (Fig.~\ref{fig:motivation_statistics}). Even when a group contains both correct and incorrect responses, vanilla GRPO broadcasts the same scalar advantage to every token in a trajectory, penalizing or rewarding a hallucinated object description no differently than surrounding boilerplate text. In short, visual errors often go uncorrected due to both vanishing gradients on hard examples and coarse credit assignment across tokens.

An alternative line of work explores on-policy distillation (OPD) for visual reasoning, using teacher predictions to provide token-level supervision. Vision-OPD~\cite{yuan2026visionopd} supervises a full-image student's generated prefixes with next-token distributions from a crop-conditioned self-teacher, following OPD~\cite{agarwal2024opd}. The cropped image fed to the teacher makes relevant details easier to recognize and provides a distributional learning signal even when sampled answers are incorrect. However, because Vision-OPD's learning objective strictly matches token-level distributions without explicitly rewarding final answer correctness, correct predictions can remain sparse. For example, in the aforementioned audit, Vision-OPD correctly answers ``How many tiers or roofs are visible on the pagoda structure?'' in only 2/8 responses %
(Fig.~\ref{fig:motivation_cases}\subref{fig:motivation_case_pagoda}). This motivates combining visual guidance from distillation with explicit outcome correctness.

The above examples suggest two complementary signals: visual distillation can supply supervision when GRPO fails, and vice versa.
Supporting the former, our audit of 1595 questions shows that on 81.87\% of the questions where none of GRPO's rollouts were correct, Vision-OPD produces at least one correct response, with examples showing object reference and fine detail perception issues with the GRPO responses.
Conversely, Vision-OPD has only 1--2 correct rollouts on 18.62\% of questions. On 292/1595 questions (18.31\%), GRPO produces at least one correct response while Vision-OPD produces at most two. This motivates using outcome rewards to increase the probability of correct responses alongside visual supervision (Sec.~\ref{sec:vrlsd_motivation}).

Building on these observations, we propose \textbf{\methodname{}} (\emph{\methodfullname{}}), a unified framework that dynamically adapts supervision based on the outcomes of the student's rollouts (Fig.~\ref{fig:overall_comparison} and Fig.~\ref{fig:visual_rlsd_framework}). For rollout groups containing both correct and incorrect responses, a %
teacher scores each sampled token under the cropped and full images. The crop-to-full token log-probability ratio redistributes GRPO credit across tokens, while answer correctness determines the reinforcement direction. We address the missing supervision in all-fail groups with an OPD fallback: a correct teacher answer activates distillation on the original student rollouts. Both branches train a student that uses standard full-image inputs at inference time.

We evaluate \methodname{} on Qwen3.5-4B and 9B across six fine-grained visual reasoning tracks and four general holdout benchmarks. At both scales, \methodname{} achieves the highest weighted fine-grained accuracy, outperforming SFT, GRPO, DAPO~\cite{yu2025dapo}, OPSD~\cite{zhao2026opsd}, and Vision-OPD. Notably, despite relying only on a same-sized teacher, \methodname{} even surpasses OPD variants distilled from a much larger 27B-parameter teacher. A naive approach of sequential Vision-OPD followed by GRPO reduces accuracy relative to Vision-OPD alone, reinforcing the need for coordinated supervision between the two methods. \methodname{} retains broad multimodal capabilities, with leading results on CV-Bench~\cite{tong2024cambrian} and POPE~\cite{li2023pope}. Case analyses connect these improvements to complementary abilities: recognizing small visual details, collecting evidence across the scene, and reasoning correctly from extracted information.
We illustrate that \methodname{} not only retains the respective strengths of GRPO and Vision-OPD on cases where one baseline fails, but can even solve some complex queries that neither baseline can handle alone.

Overall, this paper makes the following contributions:
\begin{itemize}
    \item \textbf{Analysis of complementary learning signals.} We analyze training rollouts and visual failure cases to identify missing outcome supervision in all-fail groups, coarse token credit, and unreliable success under visual distillation (Sec.~\ref{sec:vrlsd_motivation}; Fig.~\ref{fig:motivation_statistics}).
    \item \textbf{\methodname{}.} We introduce outcome-guided visual token weighting and a verified all-fail OPD fallback, combining image crop-based self-supervision with RL on the student's own responses (Sec.~\ref{sec:vrlsd_method}; Fig.~\ref{fig:visual_rlsd_framework}).
    \item \textbf{Comprehensive evaluation and analysis.} We compare training strategies across two model scales and fine-grained and holdout benchmarks (Sec.~\ref{sec:vrlsd_performance}; Tab.~\ref{tab:main_benchmarks}), and showcase how \methodname{} brings together visual grounding, scene context, and reasoning over evidence (Fig.~\ref{fig:main_showcases}).
\end{itemize}

\section{Related Work}
\label{sec:related_work}

\paragraph{Reinforcement learning for visual reasoning.}
GRPO~\cite{shao2024deepseekmath} serves as the backbone for recent multimodal RL methods such as VL-Rethinker~\cite{wang2025vlrethinker} and NoisyRollout~\cite{liu2025noisyrollout}, which focus on improving sample reuse and visual exploration. In parallel, studies in text reasoning have tackled sparse rewards via branched rollouts (BREAD~\cite{zhang2025bread}) or selective updates on high-entropy tokens~\cite{wang2025hightokenentropy}. Closer to our work, PAPO~\cite{wang2025papo}, VPPO~\cite{huang2026vppo}, and PEPO~\cite{li2026pepo} inject perceptual priors or token-level visual dependency into multimodal policy optimization. Building on this direction, \methodname{} leverages crop-to-full likelihood contrasts from a self-teacher to reshape token-level advantages, while introducing a verified distillation fallback when all rollouts fail.

\paragraph{Perception and evidence acquisition.}
To capture fine-grained details, test-time search methods such as $V^*$~\cite{wu2024vstar} and $DC^2$~\cite{wang2025hrbench} explicitly crop and retrieve relevant image regions, while agentic frameworks like DeepEyes~\cite{zheng2026deepeyes}, Pixel Reasoner~\cite{wang2025pixelreasoner}, and Thyme~\cite{zhang2025thyme} train models to interactively manipulate images. An alternative paradigm internalizes regional perception during training---as in Zooming without Zooming~\cite{wei2026zwz}---so that the model requires only a single full-image pass at inference. \methodname{} adopts this training-time internalization strategy, avoiding any multi-turn cropping overhead at test time.

\paragraph{On-policy and self-distillation.}
STaR~\cite{zelikman2022star} and ReST$^{\mathrm{EM}}$~\cite{singh2024restem} learn from successful model-generated solutions. GKD~\cite{agarwal2024opd} and MiniLLM~\cite{gu2024minillm} distill on student trajectories; subsequent work studies reward extrapolation~\cite{yang2026exopd}, divergence design~\cite{jin2026entropyopd}, and supervision reliability~\cite{hou2026uniopd}. OPSD~\cite{zhao2026opsd}, SDPO~\cite{hubotter2026sdpo}, U-OPSD~\cite{li2026uopsd}, and SDFT~\cite{shenfeld2026sdft} obtain self-supervision from privileged solutions, feedback, self-consistency, or demonstrations. Privileged distillation can also degrade long-trace reasoning~\cite{kaur2026rethinkingopsd}. Our visual self-teacher is coordinated with verified outcomes through group-dependent supervision.

\paragraph{Visual on-policy distillation.}
Two recent approaches are particularly close to ours. Vision-OPD~\cite{yuan2026visionopd} identifies a regional-to-global perception gap and matches a crop-conditioned self-teacher to a full-image student along student-generated prefixes. Its training objective requires neither answer labels nor a reward verifier. VA-OPD~\cite{liu2026vaopd} measures visual advantage through a rectified teacher log-probability difference between original and detail-degraded images. It uses this signal to reweight rollouts and separately normalize distillation losses over high- and low-advantage token groups, with the original-image teacher supplying the targets. \methodname{} draws on regional self-supervision and view-dependent token scoring, but assigns them distinct roles according to the reward group. For mixed-reward groups, a signed crop-to-full score scales GRPO advantages while preserving their outcome-derived signs. When every full-image response fails, these advantages vanish; we then enable crop-teacher OPD only if a separately generated crop-teacher answer passes verification. Distillation remains on the failed student's prefixes, and the verification response serves only to determine eligibility. Appendix~\ref{sec:extended_related_work} provides an extended discussion of these research directions.

\section{\methodname{}}
\label{sec:visual_rlsd}

In this section, we present \textbf{\methodname{}} (\emph{\methodfullname{}}). We first review OPD and GRPO (Sec.~\ref{sec:vrlsd_preliminary}), then perform an audit of their complementary strengths and limitations (Sec.~\ref{sec:vrlsd_motivation}). These observations guide the design of visual credit assignment and verified all-fail distillation (Sec.~\ref{sec:vrlsd_method}).

\subsection{Preliminaries: OPD and GRPO for Visual Reasoning}
\label{sec:vrlsd_preliminary}

Given a full image $I$, question $q$, and reference answer $a^\star$, a behavior snapshot $\pi_{\mathrm{old}}$ samples $G$ responses $\{y_i\}_{i=1}^G$. A verifier assigns $R_i=R((I,q),y_i;a^\star)\in\{0,1\}$. GRPO~\cite{shao2024deepseekmath} computes
\begin{equation}
\begin{aligned}
 A_i&=\frac{R_i-\mu_R}{\sigma_R+\varepsilon_{\mathrm{num}}},\qquad
 \mu_R=\frac{1}{G}\sum_i R_i,\\
 \sigma_R^2&=\frac{1}{G}\sum_i(R_i-\mu_R)^2.
\end{aligned}
 \label{eq:vrlsd_group_advantage}
\end{equation}
Vanilla GRPO assigns this scalar advantage $A_i$ uniformly to every token in $y_i$. Consequently, when all sampled responses receive identical rewards ($R_i\equiv0$ or $R_i\equiv1$), $A_i$ collapses to zero and the outcome-driven policy gradient vanishes. OPD~\cite{agarwal2024opd} instead matches teacher and student next-token distributions on student-generated prefixes. Vision-OPD~\cite{yuan2026visionopd} conditions the self-teacher on an evidence crop $I^+$ and the student on $I$. This supplies token-level supervision even for incorrect responses. Base loss definitions are given in Appendix~\ref{sec:method_details}.

\subsection{Motivation: Need for Correctness Plus Fine-Grained Supervision}
\label{sec:vrlsd_motivation}
\begin{figure*}[tp]
\centering
\includegraphics[width=0.92\textwidth]{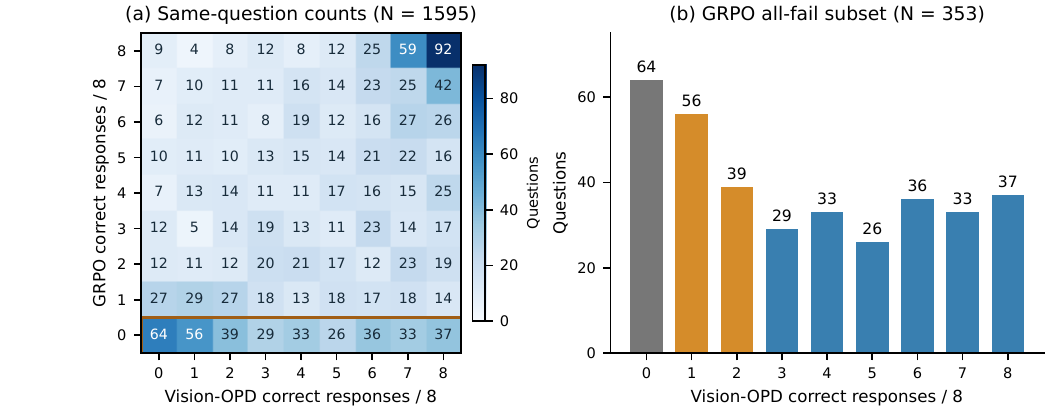}
\caption{\textbf{Complementary successes in the rollout audit (Qwen3.5-4B).} (a) Correct response counts out of 8 rollouts for 1595 matched questions; the outlined row contains 353 GRPO all-fail questions. (b) Vision-OPD counts within the GRPO all-fail subset: 289 questions have a correct rollout, including 95 questions with only 1 or 2 correct rollouts. Gray denotes 0 successes, orange 1--2, and blue 3--8.}
\label{fig:motivation_statistics}
\end{figure*}

\begin{figure*}[tp]
\centering
\begingroup
\newcommand{\auditimage}[2]{%
\begin{minipage}[c][38mm][c]{\linewidth}\centering
\includegraphics[width=\linewidth,height=38mm,keepaspectratio]{#1}%
\end{minipage}\par\vspace{1mm}{\footnotesize\textbf{#2}}%
}
\begin{subfigure}[t]{0.48\textwidth}
\centering
\begin{minipage}[t]{0.58\linewidth}\centering\auditimage{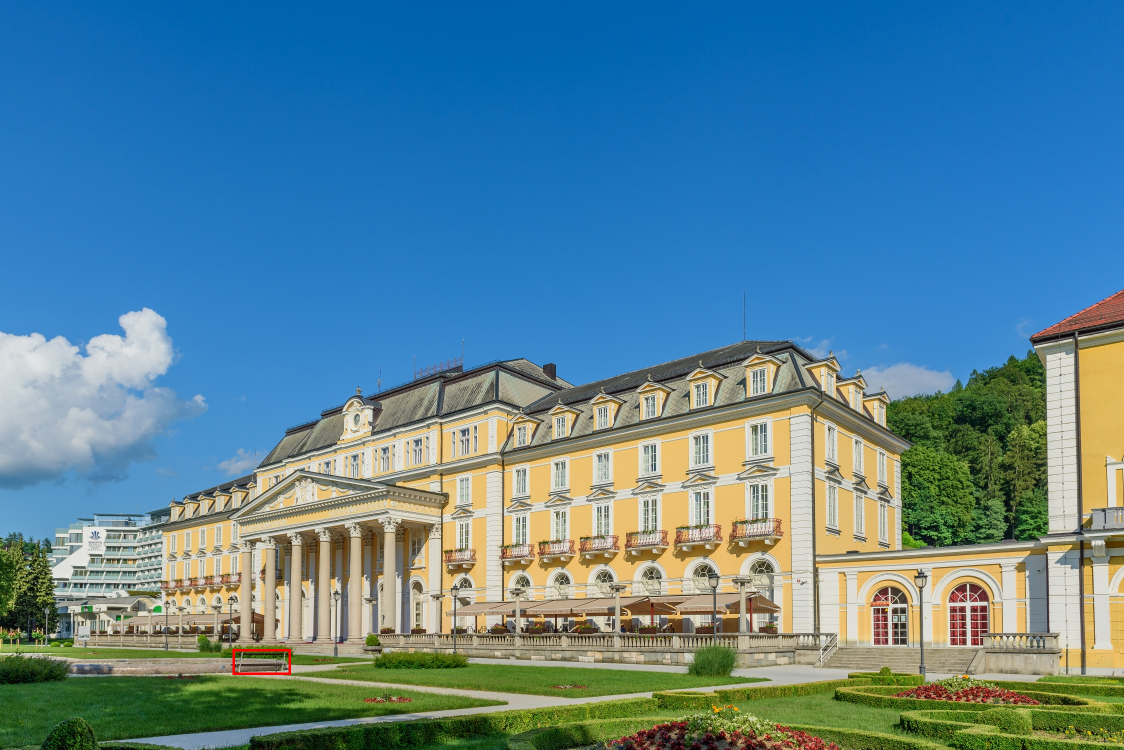}{Full image}\end{minipage}\hfill
\begin{minipage}[t]{0.38\linewidth}\centering\auditimage{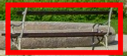}{Evidence crop}\end{minipage}
\par\smallskip\raggedright\footnotesize
\textbf{Question:} What type of object is positioned directly behind the metal bench frame?\par
\textbf{Options:} A. wooden roller; B. stone roller; C. metal bench; D. stone bench.\par
\textbf{Dataset answer:} B. stone roller.\par
\textbf{GRPO:} 0/8 correct.\par
\textbf{Vision-OPD:} 6/8 correct.\par
GRPO excerpt: ``a flat seat and backrest.''\par
\caption{Object-reference confusion (ID 2669).}
\label{fig:motivation_case_object}
\end{subfigure}
\hfill
\begin{subfigure}[t]{0.48\textwidth}
\centering
\begin{minipage}[t]{0.58\linewidth}\centering\auditimage{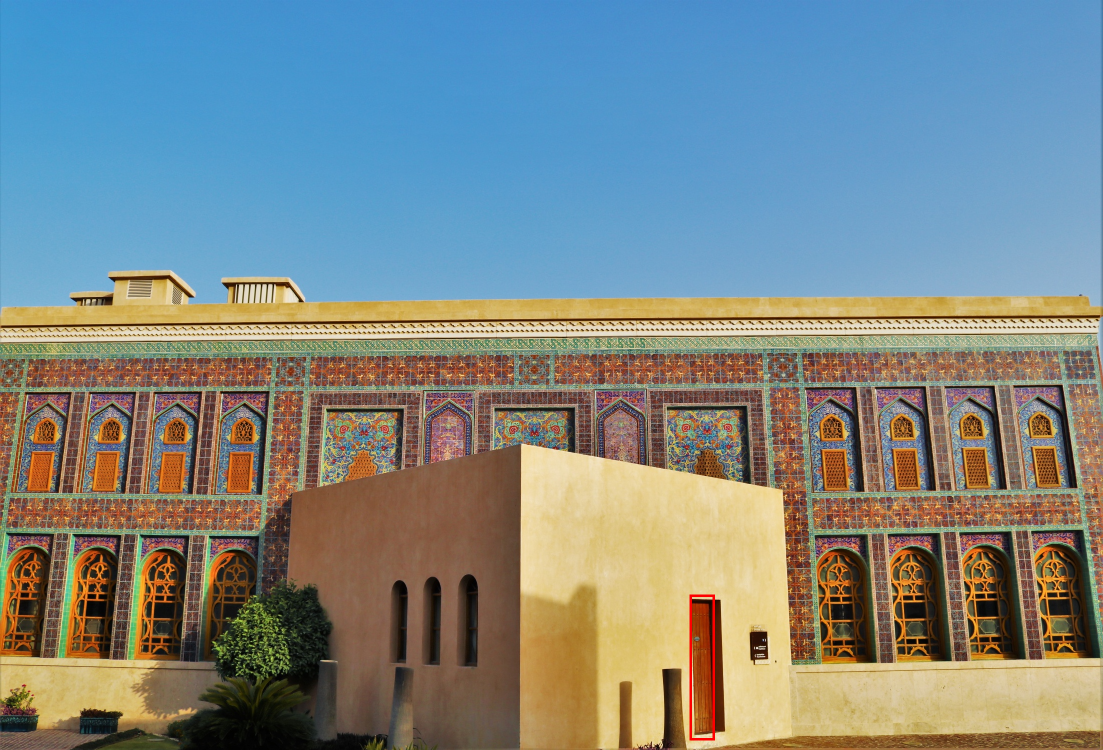}{Full image}\end{minipage}\hfill
\begin{minipage}[t]{0.38\linewidth}\centering\auditimage{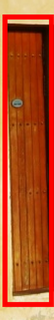}{Evidence crop}\end{minipage}
\par\smallskip\raggedright\footnotesize
\textbf{Question:} What is the shape of the metallic plate attached to the left side of the door?\par
\textbf{Options:} A. oval; B. triangular; C. circular; D. rectangular.\par
\textbf{Dataset answer:} A. oval.\par
\textbf{GRPO:} 0/8 correct.\par
\textbf{Vision-OPD:} 7/8 correct.\par
GRPO excerpt: ``four straight sides and right angles.''\par
\caption{Fine-detail shape error (ID 3092).}
\label{fig:motivation_case_plate}
\end{subfigure}
\par\medskip
\begin{subfigure}[t]{0.48\textwidth}
\centering
\begin{minipage}[t]{0.58\linewidth}\centering\auditimage{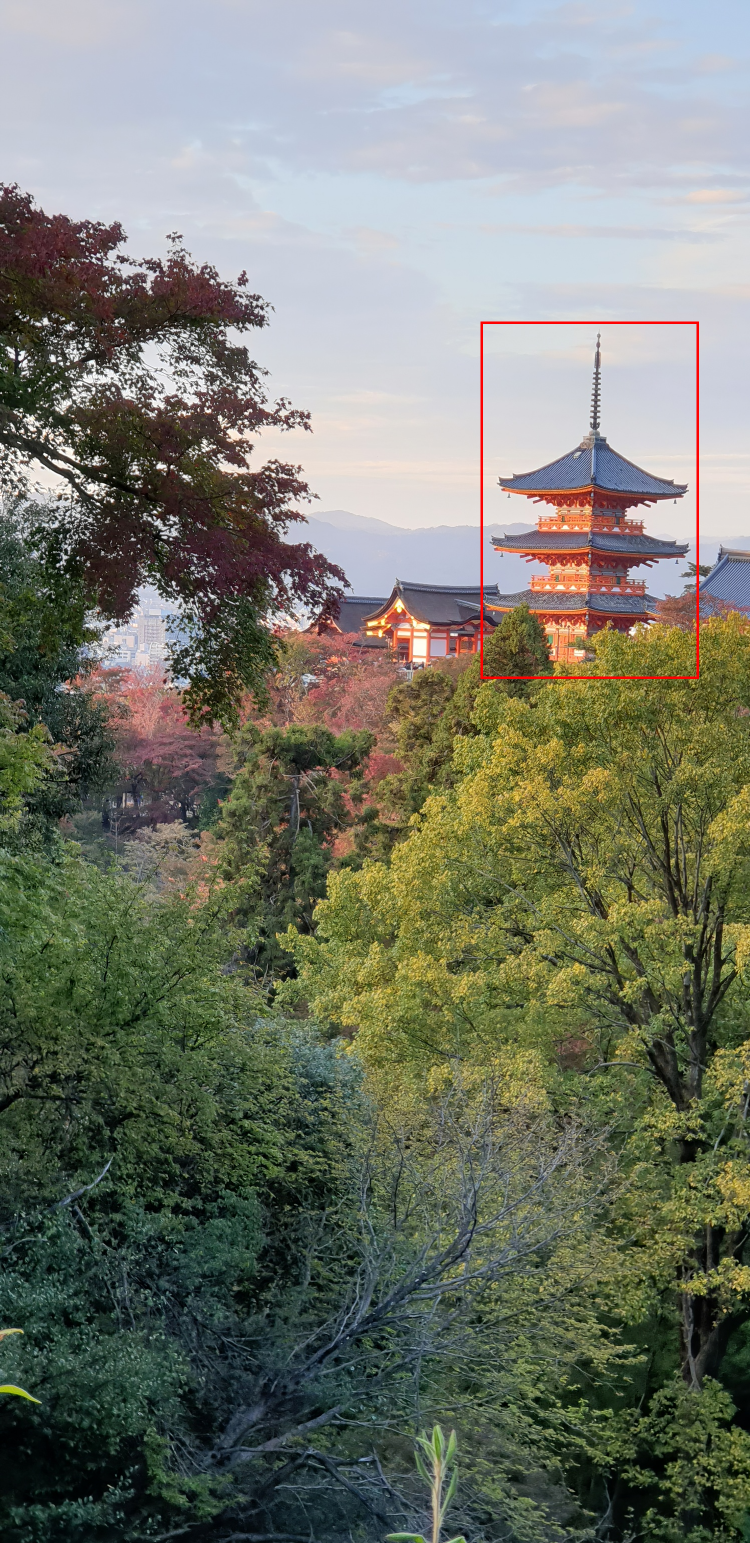}{Full image}\end{minipage}\hfill
\begin{minipage}[t]{0.38\linewidth}\centering\auditimage{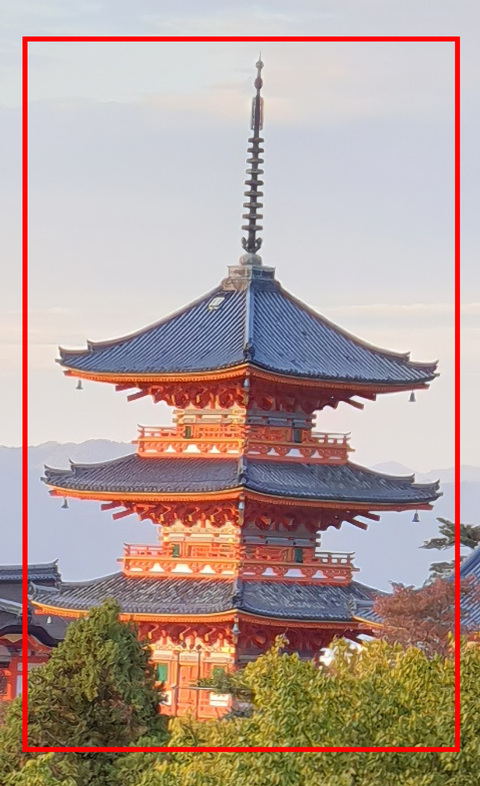}{Evidence crop}\end{minipage}
\par\smallskip\raggedright\footnotesize
\textbf{Question:} How many tiers or roofs are visible on the pagoda structure?\par
\textbf{Options:} A. 2; B. 5; C. 4; D. 3.\par
\textbf{Dataset answer:} D. 3.\par
\textbf{GRPO:} 0/8 correct.\par
\textbf{Vision-OPD:} 2/8 correct.\par
Vision-OPD answers: 3 ($2/8$ rollouts), 4 ($4/8$ rollouts), 5 ($2/8$ rollouts).\par
\caption{Correct answers remain infrequent (ID 1259).}
\label{fig:motivation_case_pagoda}
\end{subfigure}
\hfill
\begin{subfigure}[t]{0.48\textwidth}
\centering
\begin{minipage}[t]{0.58\linewidth}\centering\auditimage{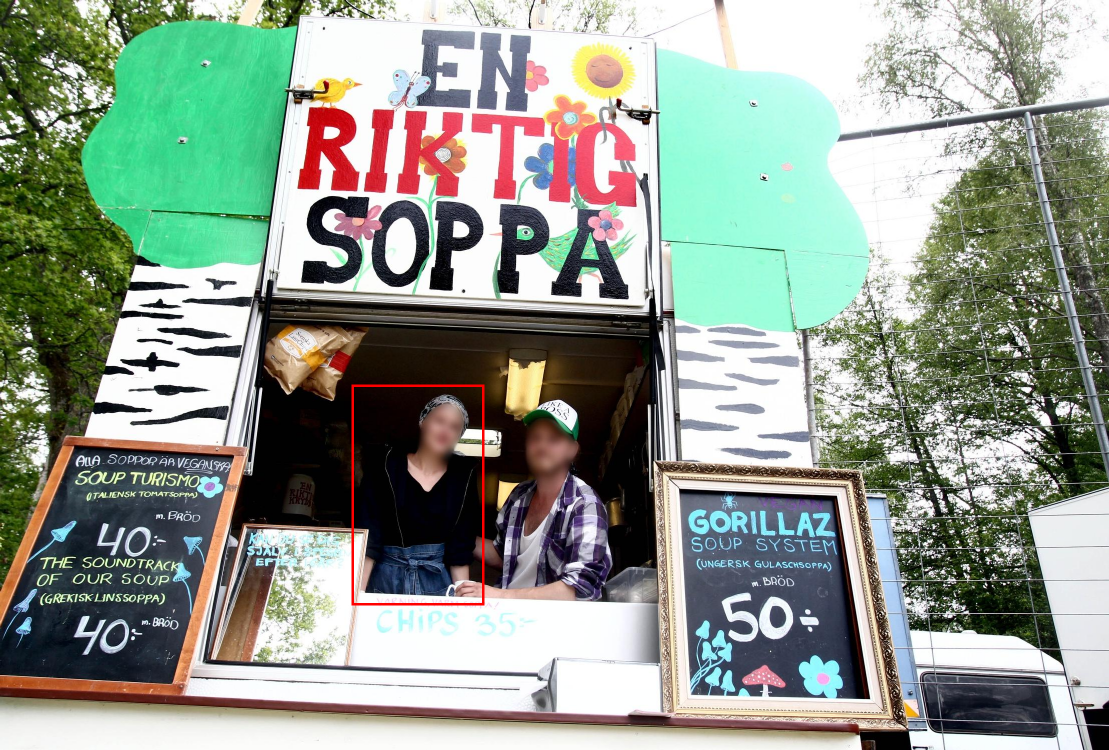}{Full image}\end{minipage}\hfill
\begin{minipage}[t]{0.38\linewidth}\centering\auditimage{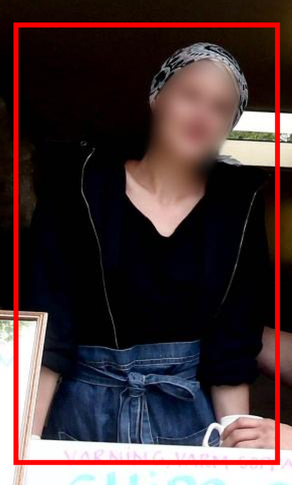}{Evidence crop}\end{minipage}
\par\smallskip\raggedright\footnotesize
\textbf{Question:} What color is the top worn underneath the dark jacket?\par
\textbf{Options:} A. gray; B. black; C. blue; D. white.\par
\textbf{Dataset answer:} B. black.\par
\textbf{GRPO:} 8/8 correct.\par
\textbf{Vision-OPD:} 0/8 correct.\par
Vision-OPD answers: white ($6/8$), blue ($1/8$), gray ($1/8$).\par
\caption{GRPO succeeds; Vision-OPD fails (ID 3035).}
\label{fig:motivation_case_clothing}
\end{subfigure}
\endgroup
\caption{\textbf{Visual failures and complementary successes in the motivating audit (\S\ref{sec:vrlsd_motivation}).} Full images and evidence crops illustrate object-reference confusion in (a), shape error in (b), sporadically correct Vision-OPD counting in (c), and a case solved by GRPO but missed by Vision-OPD in (d). Counts give the number of correct final answers out of 8 full-image rollouts. Questions target the red-box region; full details are in Appendix~\ref{sec:motivation_audit_details}.}
\label{fig:motivation_cases}
\end{figure*}

\paragraph{Audit setting.}
We randomly select 1595 samples from Vision-OPD's regional training dataset~\cite{yuan2026visionopd} and match their Qwen3.5-4B GRPO and Vision-OPD records by prompt and reference answer. Each method supplies eight full-image responses per question at temperature \samplingtemperature{}, scored with the same verifier. Let $k_G,k_O$ count correct responses for GRPO and Vision-OPD, respectively. Fig.~\ref{fig:motivation_statistics} and Fig.~\ref{fig:motivation_cases} summarize audit statistics and visual cases; full details appear in Appendix~\ref{sec:motivation_audit_details}.

\paragraph{Observation 1: GRPO and Vision-OPD exhibit complementary success patterns.}
Vision-OPD succeeds at least once on 289 questions where GRPO is all-fail ($k_G=0,k_O>0$). Conversely, 292/1595 questions (18.31\%) have $k_G\geq1$ and $k_O\leq2$ (Fig.~\ref{fig:motivation_statistics}(a)).
In Fig.~\ref{fig:motivation_cases}\subref{fig:motivation_case_clothing}, GRPO correctly answers ``black'' in all eight responses, while Vision-OPD gives ``white'', ``blue'', or ``gray'' (0/8 correct). This contrast indicates that outcome-based RL and visual distillation capture distinct, complementary aspects of visual reasoning.

\paragraph{Observation 2: When GRPO fails, Vision-OPD finds correct responses.}
GRPO produces 0/8 correct responses on 353 questions (22.13\%), leaving their outcome advantages zero. Vision-OPD finds a correct response on 289/353 of these questions (81.87\%; Fig.~\ref{fig:motivation_statistics}(b)). In Fig.~\ref{fig:motivation_cases}(a,b), GRPO describes a stone roller as a bench with a ``flat seat and backrest'' and an oval plate as having ``four straight sides and right angles''. Vision-OPD obtains 6/8 and 7/8 correct responses, respectively, against GRPO's 0/8. These failure cases highlight the need for dense visual supervision when outcome rewards alone yield zero gradient.

\paragraph{Observation 3: Vision-OPD produces only 1--2 correct responses on 18.62\% of questions.}
\phantomsection\label{obs:opd_sparse_correctness}
This occurs on 297/1595 questions. Even among the 289 questions with $k_G=0,k_O>0$, 95 (32.87\%) have only one or two correct Vision-OPD responses. In the pagoda example (Fig.~\ref{fig:motivation_cases}\subref{fig:motivation_case_pagoda}), Vision-OPD correctly counts 3 roofs only twice; four responses predict 4 roofs and two predict 5 roofs. Here, the model is capable of sampling the correct reasoning path, but does so too rarely, suggesting the need for an explicit outcome-based reward to amplify successful trajectories. %

\paragraph{Observation 4: Visual evidence provides nonuniform token-level information.}
GRPO assigns every token the same advantage $A_i$, while OPD with evidence crops changes token support unevenly. With base Qwen3.5-4B parameters and prefixes fixed, in one particular example, the crop reduces the probability of ``three'' in an incorrect response, which confuses an arched window with a triangular pediment, from $0.136$ to $4.25\times10^{-4}$. This suppression of incorrect tokens motivates redistributing outcome-derived credit on a per-token basis according to visual evidence. Full token visualizations appear in Appendix~\ref{sec:token_diagnostic_protocol} (Fig.~\ref{fig:token_importance}).

\subsection{Visual Credit Assignment with All-Fail OPD Fallback}
\label{sec:vrlsd_method}

\begin{figure*}[tp]
\centering
\includegraphics[width=0.91\textwidth]{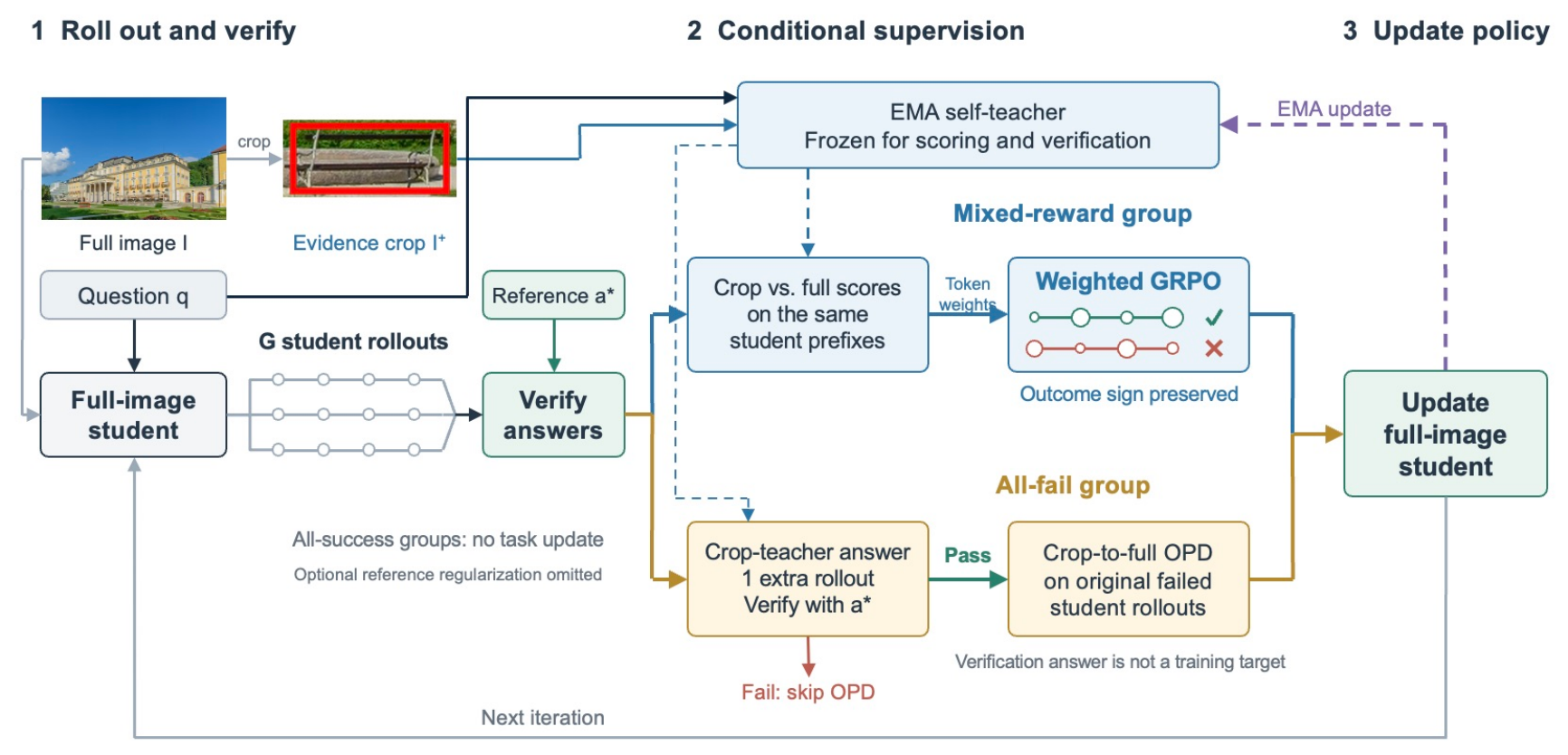}
\caption{\textbf{Training framework of \methodname{}.} (1) Verified full-image rollouts determine supervision in (2): crop-to-full scores redistribute GRPO credit in mixed-reward groups, while a verified crop-teacher answer activates OPD in all-fail groups. In (3), teacher parameters, targets, and token weights remain fixed during student optimization on the current batch. The teacher parameters are then updated by EMA for the next rollout batch. Inference uses the full-image student. Solid arrows indicate training flow; dashed arrows indicate teacher scoring and updates.}
\label{fig:visual_rlsd_framework}
\end{figure*}

These observations motivate fine-grained visual token credit for mixed-reward rollout groups and verified OPD for all-fail groups. Fig.~\ref{fig:visual_rlsd_framework} shows \methodname{}'s training framework.
Following Vision-OPD, training pairs full images, marked target regions, evidence crops, questions, and reference answers. The same-scale exponential-moving-average (EMA) self-teacher $\pi_{\bar\theta}$ is initialized from the student and frozen within each rollout group.

\paragraph{Crop-to-full token scoring.}
For each sampled token $y_{i,t}$ and its fixed prefix, the same teacher defines $p^+_{i,t}=\pi_{\bar\theta}(\cdot\mid I^+,q,y_{i,<t})$ and $p^0_{i,t}=\pi_{\bar\theta}(\cdot\mid I,q,y_{i,<t})$. We compute
\begin{equation}
 \Delta_{i,t}=\operatorname{sg}\!\left[
 \log p^+_{i,t}(y_{i,t})-\log p^0_{i,t}(y_{i,t})\right].
 \label{eq:vrlsd_scores}
\end{equation}
Using the same frozen teacher parameters $\bar{\theta}$ for both views isolates the perceptual contribution of the evidence crop, where $\operatorname{sg}[\cdot]$ denotes the stop-gradient operator. A positive $\Delta_{i,t}$ indicates that the zoomed-in crop increases the likelihood of token $y_{i,t}$, whereas a negative value indicates that the crop suppresses it.

\paragraph{Outcome-guided visual credit.}
For mixed-reward rollout groups, we convert these contrasts to positive weights, then clip and center them within each response:
\begin{equation}
\begin{aligned}
 u_{i,t}&=\exp\!\left(\operatorname{sign}(A_i)\Delta_{i,t}\right),\\
 v_{i,t}&=\operatorname{clip}(u_{i,t},1-\epsilon_w,1+\epsilon_w),\\
 w_{i,t}&=1+(v_{i,t}-\mu_i)/s_i,\qquad
 \widehat A_{i,t}=\operatorname{sg}[A_iw_{i,t}].
\end{aligned}
 \label{eq:vrlsd_credit}
\end{equation}
Here $\mu_i=T_i^{-1}\sum_t v_{i,t}$ averages over valid response tokens, and
\[
 s_i=\max\!\left\{1,\frac{\max_t|v_{i,t}-\mu_i|}{\epsilon_w}\right\}.
\]
This scaling factor $s_i$ shrinks the centered deviations only when they would otherwise exceed the clipping bounds, guaranteeing that the resulting weights $w_{i,t}$ have unit mean across the sequence and lie within $[1-\epsilon_w,1+\epsilon_w]$, where $0<\epsilon_w<1$. Weights increase with the crop-to-full token log-probability ratio $\Delta_{i,t}$ in correct responses and decrease with it in incorrect responses. Thus, visual evidence determines which tokens are emphasized and answer correctness determines the update direction. Positive weights preserve the outcome-derived sign, and centering preserves the response mean: $T_i^{-1}\sum_t\widehat A_{i,t}=A_i$. Uniform log-ratios recover the original GRPO coefficients. The PPO-style clipped loss $\ell_{\mathrm{PG}}(\widehat A)$ uses $\widehat A_{i,t}$ instead of $A_i$ (Eq.~\eqref{eq:vrlsd_pg}, Sec.~\ref{sec:policy_base_objectives}). Policy-ratio and token-weight bounds are separate.

\paragraph{Verified fallback and combined objective.}
Let $H$ indicate mixed rewards and $F$ indicate all-fail groups. Only when $F=1$, the crop teacher generates one additional answer $z$. The same verifier checks $z$ against $a^\star$, giving $C=1$ if correct and $C=0$ otherwise. Crucially, the rollout $z$ serves strictly as a gating check rather than a training target: once verified ($C=1$), the student is trained via on-policy distillation on its own failed prefixes using the teacher's next-token distributions. Specifically, our fallback loss $\ell_{\mathrm{OPD}}^+$ minimizes a sampled-token approximation of the Jensen--Shannon divergence (JSD) without token reweighting. The total objective is
\begin{equation}
\begin{aligned}
 \mathcal{L}_{\mathrm{\methodname}}
 &=\mathbb{E}\!\left[H\ell_{\mathrm{PG}}(\widehat A)
 +\lambda_{\mathrm{F}}FC\ell_{\mathrm{OPD}}^+\right]\\
 &\quad+\beta\mathcal{L}_{\mathrm{ref}}.
\end{aligned}
 \label{eq:vrlsd_total}
\end{equation}
The expectation averages over the original batch groups. $\mathcal{L}_{\mathrm{ref}}$ is optional fixed-reference KL regularization. All-success groups and failed teacher checks contribute no task loss. Teacher targets, log-ratios, weights, and verification decisions stay fixed during student updates. Each EMA update retains 95\% of the old teacher and incorporates 5\% of the current student. Inference uses the full-image student alone. Appendix~\ref{sec:method_details} gives the complete loss functions and Alg.~\ref{alg:visual_rlsd}.

\section{Experiments}
\label{sec:vrlsd_experiments}

In this section, we first describe the experimental setup (Sec.~\ref{sec:vrlsd_setup}). We then present results and analyze how \mbox{\methodname{}} combines the strengths of GRPO and Vision-OPD (Sec.~\ref{sec:vrlsd_performance}). 
Further details are in Appendices~\ref{sec:supp_evaluation} and~\ref{sec:additional_results}.
\begin{table*}[tp]
\centering
\small
\renewcommand{\arraystretch}{0.95}
\setlength{\tabcolsep}{2pt}
\newcommand{\benchmarkdashrule}{%
  \noalign{\vskip 2pt
    \hbox to \textwidth{\xleaders\hbox{\vrule width 2pt height 0.4pt depth 0pt\kern 2pt}\hfill}%
    \vskip 2pt}}
\begin{tabular*}{\textwidth}{@{\extracolsep{\fill}}lccccccccccc@{}}
\toprule
& \multicolumn{7}{c}{Fine-Grained Visual Tasks} & \multicolumn{4}{c}{Holdout Tasks} \\
\cmidrule(lr){2-8}\cmidrule(l){9-12}
& V* & Zoom & \multicolumn{2}{c}{HR-Bench} & \multicolumn{2}{c}{MME-RealWorld} & Fine & MMVP & CV- & MMStar & POPE \\
Method & Bench & Bench & 4K & 8K & EN & CN & Avg. & & Bench & & \\
\midrule
\multicolumn{12}{l}{\textit{Qwen3.5-4B}} \\
Base model & 84.29 & 47.69 & 84.38 & 80.13 & 63.70 & 63.77 & 64.34 & 76.67 & 87.13 & 78.53 & 88.28 \\
SFT on crop responses & 88.48 & 58.58 & 81.38 & 78.00 & 71.46 & 68.50 & 71.09 & \textbf{79.33} & 85.97 & 75.53 & 89.24 \\
GRPO & 83.77 & 55.38 & 82.63 & 78.25 & 66.94 & 64.66 & 66.99 & \textbf{79.33} & 87.24 & 70.60 & 86.37 \\
DAPO & 84.82 & 55.74 & 84.00 & 78.00 & 67.64 & 62.14 & 67.08 & \textbf{79.33} & 86.95 & 72.27 & 86.62 \\
OPSD & 85.34 & 53.85 & 82.25 & 79.38 & 68.20 & 69.27 & 68.75 & 78.67 & 87.27 & 75.07 & 88.83 \\
Vision-OPD & \textbf{89.01} & \textbf{59.76} & 84.50 & 80.38 & 74.88 & 70.76 & 74.18 & 78.67 & 87.11 & 78.33 & 89.11 \\
Vision-OPD+GRPO & 84.29 & 58.93 & 77.62 & 74.50 & 71.46 & 66.59 & 70.54 & 76.33 & 81.13 & 76.13 & 87.73 \\
\benchmarkdashrule
OPD (27B teacher) & 83.77 & 51.83 & 85.38 & 82.00 & 74.48 & 71.81 & 73.91 & 78.00 & 88.17 & \textbf{80.73} & 89.28 \\
VA-OPD (27B teacher) & 83.77 & 50.30 & 85.00 & 82.12 & 74.62 & 72.16 & 74.03 & 77.67 & 88.62 & 78.73 & 89.32 \\
\benchmarkdashrule
\textbf{\methodname{} (ours)} & 87.96 & 58.34 & \textbf{86.88} & \textbf{83.38} & \textbf{75.74} & \textbf{72.18} & \textbf{75.17} & 78.00 & \textbf{88.85} & 79.60 & \textbf{89.33} \\
\quad $-$fallback & 85.86 & 57.04 & 85.50 & 82.38 & 74.28 & 71.95 & 73.95 & 77.67 & 88.02 & 79.07 & 89.30 \\
\midrule
\multicolumn{12}{l}{\textit{Qwen3.5-9B}} \\
Base model & 82.72 & 52.07 & 85.75 & 80.63 & 71.40 & 67.67 & 70.86 & \textbf{83.33} & 88.29 & 83.07 & 88.88 \\
SFT on crop responses & 83.25 & 60.59 & 83.88 & 80.75 & 74.70 & 70.20 & 73.93 & 82.33 & 88.36 & 78.60 & 89.11 \\
GRPO & 85.34 & 57.51 & 86.88 & 83.25 & 73.01 & 68.50 & 72.45 & 81.67 & 87.78 & 73.40 & 87.78 \\
DAPO & 88.48 & 55.62 & 86.25 & 84.13 & 74.34 & 70.24 & 73.72 & 79.33 & 87.30 & 75.67 & 87.54 \\
OPSD & 89.53 & 57.51 & 84.00 & 81.25 & 72.61 & 70.86 & 72.49 & 80.67 & 87.45 & 79.13 & 87.48 \\
Vision-OPD & 90.05 & \textbf{64.97} & 88.13 & 85.38 & 73.40 & 70.46 & 73.40 & 82.00 & 88.27 & \textbf{83.20} & 89.13 \\
Vision-OPD+GRPO & 89.53 & 63.55 & 75.50 & 71.88 & 67.17 & 66.17 & 67.35 & 73.00 & 80.50 & 65.07 & 89.16 \\
\benchmarkdashrule
OPD (27B teacher) & 86.91 & 51.48 & 84.88 & 83.12 & 74.99 & 72.50 & 74.43 & 81.00 & 88.36 & 82.93 & 89.68 \\
VA-OPD (27B teacher) & 86.91 & 52.78 & 85.75 & 82.00 & 75.06 & 71.95 & 74.41 & 82.00 & 88.43 & 82.33 & 89.56 \\
\benchmarkdashrule
\textbf{\methodname{} (ours)} & \textbf{90.58} & 63.08 & \textbf{88.25} & \textbf{86.25} & \textbf{75.54} & \textbf{73.10} & \textbf{75.44} & 83.00 & \textbf{89.09} & 82.93 & \textbf{89.94} \\
\quad $-$fallback & 87.96 & 62.37 & 87.13 & 84.50 & 73.00 & 69.53 & 72.81 & 82.67 & 88.41 & 81.80 & 89.90 \\
\bottomrule
\end{tabular*}
\caption{\textbf{Main benchmark comparison, with \methodname{} generally outperforming other methods.} Accuracy (\%) following Vision-OPD~\cite{yuan2026visionopd}; bold marks each backbone's best scores, including ties. Fine Avg. weights the six fine-grained tracks by sample count; holdout scores are separate. Dashed groups use a 27B teacher. $-$fallback removes the all-fail OPD branch. Settings are in Appendix~\ref{sec:supp_evaluation}.}
\label{tab:main_benchmarks}
\end{table*}

\begin{figure*}[t]
\centering
\includegraphics[width=0.92\textwidth]{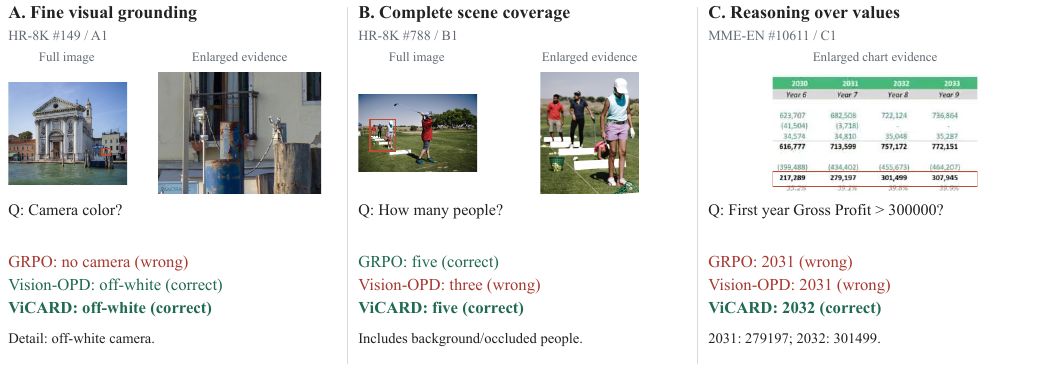}
\caption{\textbf{Complementary successes in three illustrative cases.} (A) \methodname{} and Vision-OPD recover a detail missed by GRPO. (B) \methodname{} and GRPO cover people omitted by Vision-OPD. (C) Only \methodname{} correctly associates values with years. Cropped images and bounding boxes are for display only, and are not fed into \methodname{} at inference time. Expanded views and more analyses are in Appendix~\ref{sec:additional_results}.}
\label{fig:main_showcases}
\end{figure*}

\begin{figure}[t]
\centering
\includegraphics[width=0.85\linewidth]{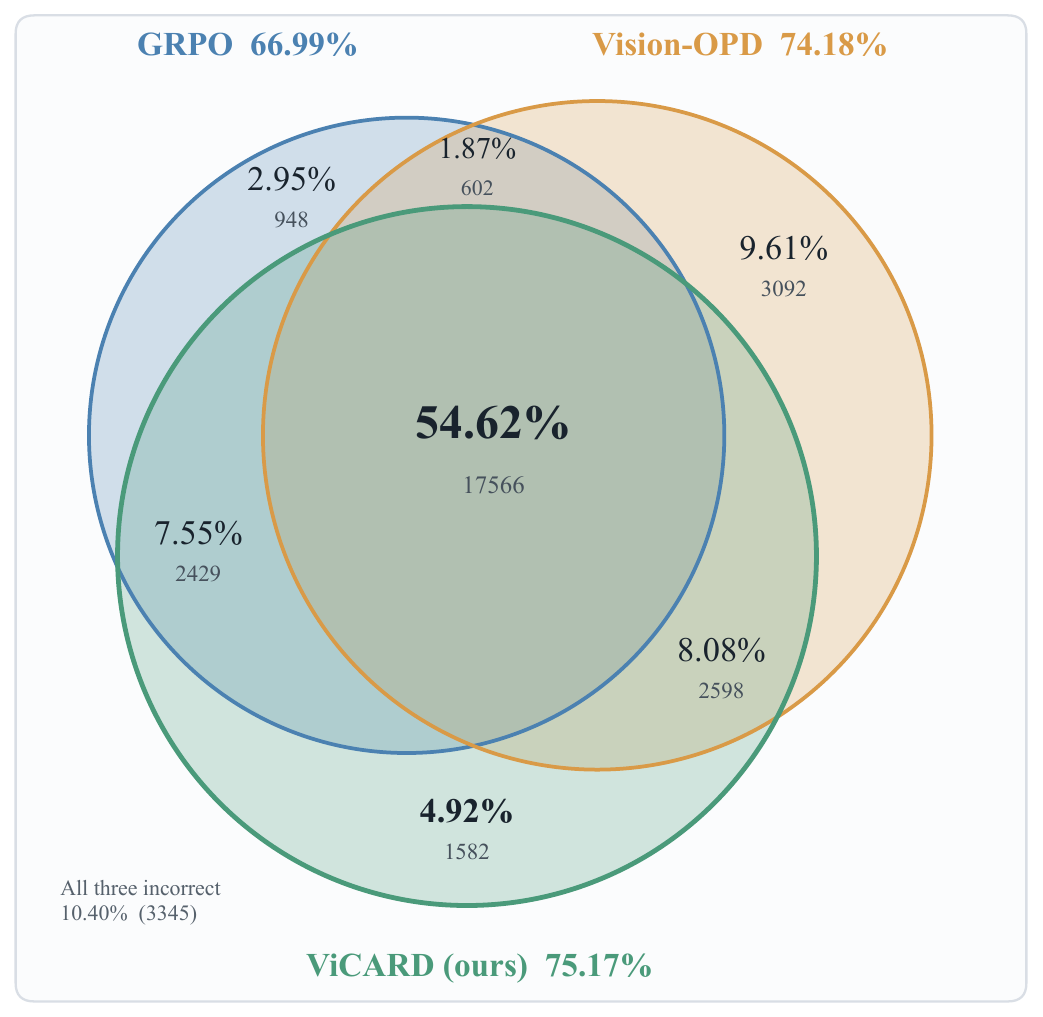}
\caption{\textbf{Correct-answer overlap between GRPO, Vision-OPD, and \methodname{} with Qwen3.5-4B.} Labels show counts and percentages of all 32162 fine-grained visual questions. \methodname{} provides complementary strengths from both baselines, capturing a large portion of the questions that GRPO and Vision-OPD can individually answer well.}
\label{fig:main_success_overlap}
\end{figure}

\subsection{Experimental Setup}
\label{sec:vrlsd_setup}

\paragraph{Data and comparisons.}
We follow Vision-OPD's regional training setting~\cite{yuan2026visionopd}, using 6241 examples with paired full images and evidence crops. All post-trained baselines use this identical question pool (see Sec.~\ref{sec:visual_view_details} for visual input formatting). We train Qwen3.5-4B and 9B~\cite{qwen2026qwen35}, comparing the base model, crop-response SFT, GRPO~\cite{shao2024deepseekmath}, DAPO~\cite{yu2025dapo}, OPSD~\cite{zhao2026opsd}, Vision-OPD, sequential Vision-OPD+GRPO, and OPD/VA-OPD with a 27B teacher. All trained methods run for one epoch; each run's best-performing checkpoint is used across benchmarks. Training settings are in Tab.~\ref{tab:experimental_setup}.

\paragraph{Evaluation.}
Following Vision-OPD, we evaluate V* Bench~\cite{wu2024vstar}, ZoomBench~\cite{wei2026zwz}, HR-Bench 4K/8K~\cite{wang2025hrbench}, and MME-RealWorld EN/CN~\cite{zhang2025mmerealworld}, plus MMVP~\cite{tong2024mmvp}, CV-Bench~\cite{tong2024cambrian}, MMStar~\cite{chen2024mmstar}, and POPE~\cite{li2023pope}. Evaluation uses non-thinking sampling (temperature \samplingtemperature{}, top-$p=\samplingtopp$, \mbox{top-$k=\samplingtopk$}) on full images. Following the official evaluation pipeline (Appendix~\ref{sec:supp_evaluation}), we report accuracy (\%) across all benchmarks, along with a sample-weighted average (\textbf{Fine Avg.}) over the six fine-grained tracks (32162 questions in total), with the four general holdout scores reported separately.

\subsection{Main Benchmark Results}
\label{sec:vrlsd_performance}

\paragraph{Stronger fine-grained performance across model scales.}
\methodname{} achieves the highest Fine Avg. across both backbones, reaching 75.17\% with the 4B backbone and 75.44\% for 9B (+10.83 and +4.58 points over the respective base models; Tab.~\ref{tab:main_benchmarks}). Crucially, it outperforms outcome-only RL (GRPO, DAPO) and answer-conditioned OPSD on every fine-grained track at both scales, with aggregate gains over GRPO reaching 8.18 (4B) and 2.99 points (9B). This consistent margin indicates that neither outcome verification nor privileged text answers alone can resolve fine-grained perceptual bottlenecks; injecting localized visual supervision directly into policy optimization is needed to ground small visual details.

\paragraph{Further gains over visual distillation.}
\methodname{} surpasses Vision-OPD by 0.99 (4B) and 2.04 (9B) Fine Avg. points, driven by consistent improvements on high-resolution (HR-Bench 4K/8K) and complex real-world scenes (MME-RealWorld EN/CN)---tasks where distribution matching alone often yields sporadic correct answers (\hyperref[obs:opd_sparse_correctness]{Observation~3}). While Vision-OPD retains an edge on ZoomBench and 4B $V^*$ Bench, \methodname{} takes the lead on 9B $V^*$ Bench. More importantly, despite relying solely on a same-sized EMA self-teacher, \methodname{} outperforms both 27B-teacher distillation baselines (OPD and VA-OPD) across all six tracks. This suggests that fine-grained visual bottlenecks stem primarily from resolution and grounding gaps rather than model capacity: pairing a cropped view with outcome-verified credit assignment provides a more targeted learning signal than distilling from a much larger full-image teacher, while eliminating 27B-scale inference overhead.

\paragraph{Sequential distillation and RL can degrade performance.}
A naive approach of applying GRPO after Vision-OPD reduces Fine Avg. from 74.18\% to 70.54\% at 4B and from 73.40\% to 67.35\% at 9B: drops of 3.64 and 6.05 points. All six tracks decline at both scales, and 9B MMStar falls from 83.20\% to 65.07\%. We hypothesize that switching between crop-conditioned distribution matching and outcome-only RL, without coordinating their update directions, disrupts learned visual behavior and causes performance collapse. \methodname{} instead integrates visual guidance seamlessly into outcome optimization.

\paragraph{Retaining general multimodal capabilities.}
These capabilities span visual understanding, spatial reasoning, and resistance to object hallucination. GRPO and DAPO improve fine-grained accuracy but degrade these capabilities, reducing MMStar accuracy by up to 7.93 (4B) and 9.67 points (9B) and lowering POPE scores relative to the base models. We attribute this degradation to coarse sequence-level credit assignment, which indiscriminately reinforces all tokens in successful regional rollouts and overfits the policy to narrow task priors. By contrast, \methodname{}'s sequence-centered visual weighting localizes credit adjustments to visually decisive tokens while keeping mean response advantages unchanged. Consequently, \methodname{} achieves the best overall performance on CV-Bench and POPE at both scales, recovers 9.00--9.53 points over GRPO on MMStar, and broadly matches or improves upon the base models across holdout tasks.

\paragraph{Contribution of verified fallback.}
Ablating the all-fail OPD fallback (\texttt{-fallback} in Tab.~\ref{tab:main_benchmarks}) reduces Fine Avg. by 1.22 points at 4B (to 73.95\%) and 2.63 points at 9B (to 72.81\%). Notably, without the fallback, the 9B model's performance drops nearly to the level of vanilla GRPO (72.45\%). This confirms that even as backbone capacity scales, hard perceptual queries still trigger all-fail rollout groups where outcome gradients vanish; rescuing these wasted rollouts via verified crop-teacher distillation is critical to reach the full potential of larger models.

\paragraph{Combining complementary successes.}
Fig.~\ref{fig:main_success_overlap} compares the correct-answer sets using the 4B backbone model. The areas of the overlap roughly correspond to the labeled percentages. \methodname{} answers 2598 questions missed by GRPO but solved by Vision-OPD (8.08\%), and 2429 missed by Vision-OPD but solved by GRPO (7.55\%). It additionally answers 1582 questions missed by both (4.92\%). This breakdown suggests that \methodname{} successfully consolidates the complementary strengths of both baselines while solving problems where both fail.

\paragraph{Illustrative examples.}
Fig.~\ref{fig:main_showcases} illustrates one representative case from each overlap regime in Fig.~\ref{fig:main_success_overlap} (see Appendix~\ref{sec:additional_results} and Fig.~\ref{fig:additional_showcases} for all six cases):
\begin{list}{\textbullet}{%
  \setlength{\leftmargin}{0.9em}%
  \setlength{\labelwidth}{0.5em}%
  \setlength{\labelsep}{0.4em}%
  \setlength{\topsep}{1pt}%
  \setlength{\itemsep}{1pt}%
  \setlength{\parsep}{0pt}%
  \setlength{\partopsep}{0pt}}
\item \emph{Recovering fine details (Group A):} In A1 (Fig.~\ref{fig:main_showcases} left), GRPO fails to spot a tiny off-white surveillance camera in an 8K scene, whereas both Vision-OPD and \methodname{} identify it correctly. In A2, GRPO locates the right table row (Year~7) but miscalculates the calendar year as 2029 instead of $2025+(7-1)=2031$. Here, crop-guided supervision helps \methodname{} anchor both fine-grained perception and downstream arithmetic.
\item \emph{Avoiding local-crop bias (Group B):} Conversely, pure crop-based distillation can induce tunnel vision or neglect global question constraints. In B1 (Fig.~\ref{fig:main_showcases} center), Vision-OPD counts only three foreground figures, missing a background person in red and an occluded person in pink; in B2, it blindly reports the chart's grand total (228M USD) rather than summing only the two requested assets ($85+71=156\text{M}$ USD). Because \methodname{} optimizes outcome rewards over full-image rollouts, it preserves global scene coverage and strict adherence to question scope.
\item \emph{Synergistic binding and reasoning (Group C):} In C1 (Fig.~\ref{fig:main_showcases} right) and C3, both baselines fail on dense financial charts. In C1, GRPO shifts values into the empty START/Year~0 column.
By jointly reinforcing correct final outcomes and weighting visually grounded tokens, \methodname{} accurately maps fine-grained visual values to their structural labels.
\end{list}

\section{Conclusions}
\label{sec:conclusion}

We presented \methodname{}, which combines visual self-distillation with outcome-based RL. Motivated by the complementary limitations of GRPO and Vision-OPD, it refines token credit using visual evidence and supplies verified distillation for all-fail groups. Experiments across model scales show stronger fine-grained performance while retaining general multimodal capabilities, with complementary gains in detail recognition, scene coverage, and reasoning. These results support coordinating visual supervision and outcome feedback for reliable multimodal reasoning.

\label{page:main_conclusion_end}

\FloatBarrier
\raggedbottom
\captionsetup[figure]{skip=10pt}
\setlength{\dbltextfloatsep}{6pt plus 2pt minus 2pt}
\section*{LLM Usage Statement}
OpenAI's ChatGPT and Codex assisted with language polishing. Codex and Claude Code also assisted with coding. The authors are responsible for the content.

{
    \small
    \bibliographystyle{ieeenat_fullname}
    \bibliography{main}
}

\appendix
\section*{Appendix}
The supplementary material is organized as follows:
\begin{itemize}
    \item \textbf{Appendix~\ref{sec:method_details}} gives the complete training algorithm and objective definitions, including token weighting, sequence centering, and verified OPD fallback.
    \item \textbf{Appendix~\ref{sec:supp_evaluation}} documents data construction, evaluation protocols, baseline conditions, and training settings.
    \item \textbf{Appendix~\ref{sec:motivation_audit_details}} details the matched rollout audit, answer verification, and sensitivity analyses underlying the motivation observations.
    \item \textbf{Appendix~\ref{sec:additional_results}} provides additional performance plots and complete benchmark case studies.
    \item \textbf{Appendix~\ref{sec:token_diagnostic_protocol}} presents paired token-importance examples, scoring and rendering protocols, and token log-ratio distributions before and after training.
    \item \textbf{Appendix~\ref{sec:on_policy_diagnostics}} examines rollout/actor agreement and per-update probability changes to assess on-policy optimization.
    \item \textbf{Appendix~\ref{sec:extended_related_work}} expands the discussion of visual reinforcement learning, visual evidence acquisition, and on-policy and self-distillation.
\end{itemize}

\section{Full Training Procedure and Objective Details}
\label{sec:method_details}
Fig.~\ref{fig:visual_rlsd_framework} shows the training loop, and Algorithm~\ref{alg:visual_rlsd} gives the complete procedure. This appendix section expands the base objectives, visual inputs, centering implementation, and verified distillation branch used in Sec.~\ref{sec:vrlsd_method}.
\begin{algorithm}[!htbp]
\caption{\methodname{} training}
\label{alg:visual_rlsd}
\footnotesize
\begin{algorithmic}[1]
\Require Data $\mathcal{D}=\{(I,I^+,q,a^\star)\}$; student $\pi_\theta$;
\Statex \hspace{\algorithmicindent} verifier $R$; optional fixed reference $\pi_{\mathrm{ref}}$;
\Statex \hspace{\algorithmicindent} group size $G$; iterations $K$; updates per batch $U$;
\Statex \hspace{\algorithmicindent} $\alpha$, $\epsilon_w$, $\epsilon_{\mathrm{P}}$, $\lambda_{\mathrm{F}}$, and $\beta$.
\Ensure Full-image policy $\pi_\theta$.
\State Initialize EMA teacher $\bar\theta\gets\theta$.
\For{$k=1,\ldots,K$}
    \State Set $\theta_{\mathrm{old}}\gets\operatorname{sg}[\theta]$; freeze $\bar\theta$.
    \State Sample a batch $\mathcal{B}\subset\mathcal{D}$.
    \For{each example $b=(I,I^+,q,a^\star)\in\mathcal{B}$}
        \State Sample $\mathcal{Y}_b=\{y_i\}_{i=1}^{G}$ from $\pi_{\mathrm{old}}(\cdot\mid I,q)$.
        \State Verify $R_i\gets R((I,q),y_i;a^\star)$ for every $i$.
        \State Compute $A_i$ by Eq.~\eqref{eq:vrlsd_group_advantage}.
        \State Compute $F_b,H_b$ by Eq.~\eqref{eq:vrlsd_group_types}; set $C_b\gets0$.
        \If{$H_b=1$} \Comment{Mixed rewards}
            \State Score $\mathcal{Y}_b$ under crop $p^+$ and full $p^0$
            \Statex \hspace{\dimexpr\algorithmicindent*3\relax} using the same frozen teacher, Eq.~\eqref{eq:vrlsd_views}.
            \State Compute detached $\Delta_{i,t}$ by Eq.~\eqref{eq:vrlsd_scores}.
            \State Clip sign-aware weights; center within each response
            \Statex \hspace{\dimexpr\algorithmicindent*3\relax} to mean one while retaining the weight bounds.
            \State Cache detached $\widehat A_b$ using Eq.~\eqref{eq:vrlsd_credit}.
        \ElsIf{$F_b=1$} \Comment{All student rollouts fail}
            \State Generate one $z\sim\pi_{\bar\theta}(\cdot\mid I^+,q)$.
            \State Set $C_b\gets R((I^+,q),z;a^\star)$.
            \If{$C_b=1$}
                \State Cache detached crop-teacher token probabilities
                \Statex \hspace{\dimexpr\algorithmicindent*4\relax} on the original student prefixes in $\mathcal{Y}_b$.
            \EndIf
        \EndIf
    \EndFor
    \For{$u=1,\ldots,U$}
        \State Evaluate $p_\theta$ on full-image student prefixes.
        \State For $H_b=1$, compute $\rho_{i,t}(\theta)$ against $\pi_{\mathrm{old}}$
        \Statex \hspace{\dimexpr\algorithmicindent*2\relax} and the clipped policy loss $\ell_{\mathrm{PG}}(\widehat A_b)$.
        \State For $F_bC_b=1$, compute sampled-token JSD loss
        \Statex \hspace{\dimexpr\algorithmicindent*2\relax} $\ell_{\mathrm{OPD},b}^+$ using cached teacher probabilities.
        \State Weight fallback OPD by $\lambda_{\mathrm{F}}$; average active losses
        \Statex \hspace{\dimexpr\algorithmicindent*2\relax} over the original $|\mathcal{B}|$ groups and add $\beta\mathcal{L}_{\mathrm{ref}}$,
        \Statex \hspace{\dimexpr\algorithmicindent*2\relax} as in Eq.~\eqref{eq:vrlsd_total}.
        \State Update $\theta$ by minimizing this loss; only the student
        \Statex \hspace{\dimexpr\algorithmicindent*2\relax} receives gradients. Keep cached targets and weights fixed.
    \EndFor
    \State Update $\bar\theta\gets(1-\alpha)\bar\theta+\alpha\theta$.
\EndFor
\State \Return $\pi_\theta(\cdot\mid I,q)$.
\end{algorithmic}
\medskip
\textit{Branch handling.} All-success groups and all-fail groups with $C_b=0$ contribute only optional reference regularization. The verified response $z$ determines fallback eligibility; crop-teacher probabilities on the original student tokens and prefixes supply the distillation targets.
\end{algorithm}

\subsection{Training Workflow}
\begin{enumerate}

    \item \textbf{Roll out and verify.} Given a full image and question, the student samples a group of $G$ responses. An answer verifier compares their final answers with $a^\star$ to determine the group outcomes.
    \item \textbf{Select conditional supervision.} In mixed-reward groups, the same EMA self-teacher scores each student prefix under the crop and full image. Their score difference determines positive token weights, clipped and centered to have mean one within each response. These weights redistribute GRPO credit while preserving the outcome advantages' signs and response means. In all-fail groups, the crop teacher generates one additional response. A correct teacher answer activates OPD on the original failed student rollouts. All-success groups and all-fail groups with an unsuccessful teacher check contribute no task loss.
    \item \textbf{Update the policy.} Optimize the full-image student using the selected losses, then refresh the EMA teacher and behavior snapshot for the next iteration. Teacher parameters remain fixed within each rollout batch. Optional reference regularization is omitted from the figure; inference uses only the full-image student.
\end{enumerate}

\subsection{Policy and Base Objectives}
\label{sec:policy_base_objectives}
\paragraph{Problem setup.}
Let $x=(I,q)$ denote an image--question pair and $a^\star$ its reference answer. A vision--language policy generates a response $y=(y_1,\ldots,y_T)$, containing a final answer and optionally intermediate reasoning, according to
\begin{equation}
    \pi_\theta(y\mid x)
    =\prod_{t=1}^{T}\pi_\theta(y_t\mid I,q,y_{<t}).
    \label{eq:vrlsd_policy}
\end{equation}
A verifier assigns the binary outcome reward $R(x,y;a^\star)\in\{0,1\}$. The reference answer is used exclusively for outcome verification. At each training iteration, a fixed behavior snapshot $\pi_{\mathrm{old}}$ generates $G$ responses $\mathcal{Y}=\{y_i\}_{i=1}^{G}$ from the full image. We write $T_i=|y_i|$ and $p_{\theta,i,t}(\cdot)=\pi_\theta(\cdot\mid I,q,y_{i,<t})$. All token sums below include generated response tokens and exclude prompt and padding tokens. Here $I$ denotes the global model input; training-time visual views are detailed in Sec.~\ref{sec:visual_view_details}.

\paragraph{On-policy distillation.}
OPD evaluates a teacher on prefixes generated by the student, so supervision addresses the states visited by the student itself~\cite{agarwal2024opd}. Let $p_{T,i,t}$ denote the teacher's next-token distribution at prefix $y_{i,<t}$. The per-group distillation loss is
\begin{equation}
    \ell_{\mathrm{OPD}}(x,\mathcal{Y})
    =\frac{1}{G}\sum_{i=1}^{G}\frac{1}{T_i}
    \sum_{t=1}^{T_i}
    D\!\left(p_{T,i,t},p_{\theta,i,t}\right).
    \label{eq:vrlsd_opd}
\end{equation}
Here $D$ is a distribution divergence over the vocabulary, with gradients stopped through the teacher and sampled prefixes. In visual self-distillation, the teacher can be a snapshot of the same model conditioned on an evidence-centered crop, while the student observes the full image~\cite{yuan2026visionopd}. The crop gives the shared model a complementary view of the visual evidence. OPD provides distributional supervision without requiring the sampled response to be correct, but the usefulness of this supervision depends on the teacher's reliability under the student's prefix.

\paragraph{GRPO for visual reasoning.}
GRPO~\cite{shao2024deepseekmath} compares the rewards $R_i=R(x,y_i;a^\star)$ within each group. We use the population-normalized advantages in Eq.~\eqref{eq:vrlsd_group_advantage}, with $\varepsilon_{\mathrm{num}}>0$ for numerical stability. With the token-level importance ratio
\begin{equation}
    \rho_{i,t}(\theta)
    =\frac{\pi_\theta(y_{i,t}\mid I,q,y_{i,<t})}
    {\pi_{\mathrm{old}}(y_{i,t}\mid I,q,y_{i,<t})},
    \label{eq:vrlsd_ratio}
\end{equation}
we define the PPO-style clipped policy loss~\cite{schulman2017ppo} for arbitrary detached coefficients $B_{i,t}$ as
\begin{equation}
\begin{aligned}
 \ell_{\mathrm{PG}}(B)
 &=-\frac{1}{G}\sum_{i=1}^{G}\frac{1}{T_i}\sum_{t=1}^{T_i}
      \min\!\bigl\{\rho_{i,t}B_{i,t},\\[-2pt]
 &\hspace{35mm}\bar\rho_{i,t}B_{i,t}\bigr\},\\
 \bar\rho_{i,t}
 &=\operatorname{clip}(\rho_{i,t},1-\epsilon_{\mathrm{P}},1+\epsilon_{\mathrm{P}}).
\end{aligned}
    \label{eq:vrlsd_pg}
\end{equation}
Vanilla GRPO uses $B_{i,t}=A_i$ at every position. Every token in a response receives the same outcome advantage; its gradient also depends on the token likelihood and policy-ratio clipping. An optional fixed-reference KL regularizer is separate from this policy loss. When all rewards are identical, $A_i=0$ for every response: the outcome-driven policy term vanishes.

\subsection{Visual Views and Conditional Optimization}
\label{sec:visual_view_details}
\paragraph{Visual views and training examples.}
We follow the regional data construction of Vision-OPD~\cite{yuan2026visionopd}. Object identification and segmentation propose small regions in raw images; a question generator produces questions answerable from each retained region. The global input $I$ contains the region's bounding-box overlay, and $q$ includes a spatial restriction to that region. The positive view $I^+$ is the corresponding crop resized by $2\times$. Reference answers follow the crop-based answer-consensus procedure used for its RLVR comparison. Thus, $\mathcal{D}$ provides $(I,I^+,q,a^\star)$, pairing the full image with a single privileged crop for policy training.

\paragraph{Crop-to-full visual scoring.}
We maintain an exponential-moving-average self-teacher with parameters $\bar\theta$, initialized from the student. At the start of an iteration, its parameters are frozen for scoring and verification on that iteration's rollout batch. For every student prefix, it defines
\begin{equation}
\begin{aligned}
 p^+_{i,t}(\cdot)&=\pi_{\bar\theta}(\cdot\mid I^+,q,y_{i,<t}),\\
 p^0_{i,t}(\cdot)&=\pi_{\bar\theta}(\cdot\mid I,q,y_{i,<t}).
\end{aligned}
    \label{eq:vrlsd_views}
\end{equation}
These distributions define the detached crop-to-full token log-ratio $\Delta_{i,t}$ in Eq.~\eqref{eq:vrlsd_scores}.
The full-image scoring distribution $p^0$ uses the same teacher parameters as $p^+$, separating view differences from teacher--student parameter drift. It is distinct from both the trainable student distribution $p_\theta$ and the behavior distribution in Eq.~\eqref{eq:vrlsd_ratio}. Positive $\Delta_{i,t}$ indicates increased support under the crop, while negative $\Delta_{i,t}$ indicates reduced support. These token log-ratios are combined with verifier-derived outcome advantages in the following step.

\paragraph{Outcome-guided visual credit.}
In mixed-reward groups, we convert crop-to-full token log-ratios into positive token weights conditioned on the outcome-advantage sign. We clip and sequence-center these weights within each response as in Eq.~\eqref{eq:vrlsd_credit}.
Here $\mu_i=T_i^{-1}\sum_t v_{i,t}$ is the mean clipped weight over valid response tokens. To retain the weight bounds, we use
\[
 s_i=\max\!\left\{1,\frac{\max_t|v_{i,t}-\mu_i|}{\epsilon_w}\right\}.
\]
This contracts the centered deviations only when needed, giving $T_i^{-1}\sum_t w_{i,t}=1$ and $w_{i,t}\in[1-\epsilon_w,1+\epsilon_w]$.

Centering preserves the ordering of the clipped weights. Subject to clipping, weights increase with $\Delta_{i,t}$ in successful responses and decrease with it in failed responses. Visual evidence thus redistributes credit within a response: $w_{i,t}>1$ increases the coefficient magnitude, and $w_{i,t}<1$ reduces it. Uniform log-ratios recover $w_{i,t}=1$ at every position. Since $w_{i,t}>0$ and its response mean is one, the reweighted coefficients preserve both the outcome-derived sign and the response mean: $T_i^{-1}\sum_t\widehat A_{i,t}=A_i$. The evidence-weight bound $\epsilon_w$ is separate from the policy-ratio clipping parameter $\epsilon_{\mathrm{P}}$.

\paragraph{Verified OPD fallback for all-fail groups.}
Define the all-fail and mixed-reward indicators
\begin{equation}
\begin{aligned}
 F&=\mathbf{1}\!\left\{\sum_{i=1}^{G}R_i=0\right\},\\
 H&=\mathbf{1}\!\left\{0<\sum_{i=1}^{G}R_i<G\right\}.
\end{aligned}
    \label{eq:vrlsd_group_types}
\end{equation}
Only when $F=1$, we generate one additional answer trajectory from the crop teacher,
\begin{equation}
 z\sim\pi_{\bar\theta}(\cdot\mid I^+,q),\qquad
 C=R((I^+,q),z;a^\star).
    \label{eq:vrlsd_teacher_check}
\end{equation}
We use the same answer verifier for $z$ and the student responses, and set $C=0$ for other group types by convention. If $C=1$, the fallback applies Eq.~\eqref{eq:vrlsd_opd} with $p_T=p^+$ to the original failed student rollouts. The verified trajectory $z$ determines the eligibility indicator $C$. Distillation targets come from the crop teacher conditioned on $(I^+,q)$ and the original student prefixes.

The fallback targets the symmetric Jensen--Shannon divergence (JSD),
\begin{equation}
\begin{aligned}
 m_{i,t}&=\tfrac12(p^+_{i,t}+p_{\theta,i,t}),\\
 D_{\mathrm{JS}}(p^+_{i,t},p_{\theta,i,t})
 &=\tfrac12 D_{\mathrm{KL}}(p^+_{i,t}\|m_{i,t})\\
 &\quad+\tfrac12 D_{\mathrm{KL}}(p_{\theta,i,t}\|m_{i,t}).
\end{aligned}
    \label{eq:vrlsd_jsd}
\end{equation}
We implement this objective with a \emph{sampled-token JSD approximation}, evaluated on the tokens in the original student rollouts. The crop teacher supplies detached probabilities at the same student prefixes, and only the student receives gradients. Each valid response position contributes to the fallback without the visual weights in Eq.~\eqref{eq:vrlsd_credit}. This supplies a distillation signal for groups with $A_i=0$. When teacher verification fails, the group contributes no fallback loss. All-success groups also receive no fallback.

\paragraph{Combined objective and training.}
Let $\ell_{\mathrm{OPD}}^+$ denote the crop-teacher loss in Eq.~\eqref{eq:vrlsd_opd}, using the sampled-token approximation to Eq.~\eqref{eq:vrlsd_jsd}. The complete objective is Eq.~\eqref{eq:vrlsd_total}, where $\lambda_{\mathrm{F}}>0$ controls fallback strength. The expectation covers training examples, full-image rollout groups, and the conditional verification rollout. The normalization denominator is the original number of groups in the batch. The optional $\mathcal{L}_{\mathrm{ref}}$ is the average tokenwise $D_{\mathrm{KL}}(p_\theta\|\pi_{\mathrm{ref}})$ on full-image student prefixes for a fixed reference policy; $\beta=0$ disables it. The reference term applies to all groups when $\beta>0$.

Each iteration samples full-image groups and verifies their answers. Mixed groups use crop-to-full token log-ratios and Eq.~\eqref{eq:vrlsd_credit}; all-fail groups invoke Eq.~\eqref{eq:vrlsd_teacher_check} and distill only if the teacher passes. After optimizing the student, we update the self-teacher as $\bar\theta\leftarrow(1-\alpha)\bar\theta+\alpha\theta$, where $\alpha=0.05$ is the student update rate: each update retains 95\% of the previous teacher and incorporates 5\% of the current student. We then refresh the behavior snapshot for the next iteration. Token log-ratios, weights, verification decisions, and teacher targets remain fixed during updates on the same rollout batch. Inference runs the full-image student on the benchmark's supplied inputs.

The two branches have different sources of supervision. In mixed groups, visual evidence changes the magnitude of verifier-derived advantage coefficients. In verified all-fail groups, the teacher distribution determines the distillation update.

\section{Data, Evaluation, and Comparison Details}
\label{sec:supp_evaluation}
\paragraph{Regional data construction.}
We follow the construction described in Sec.~3.2 of Vision-OPD~\cite{yuan2026visionopd}. Small regions are proposed through object identification and segmentation. Qwen3.5-397B generates region-answerable questions and crop-conditioned candidate answers; questions are retained when answer consensus exceeds $0.75$. The paired visual inputs are detailed in Sec.~\ref{sec:visual_view_details}.

\paragraph{Evaluation prompts and decoding.}
We use the benchmark-specific prompts in Appendix~B of Vision-OPD~\cite{yuan2026visionopd} and its released evaluation pipeline.\footnote{\url{https://github.com/VisionOPD/Vision-OPD/tree/main/eval}}
Each question receives one response with thinking disabled, using temperature \samplingtemperature{}, top-$p=\samplingtopp$, and top-$k=\samplingtopk$, following Qwen3.5's non-thinking generation recommendations.\footnote{\url{https://huggingface.co/Qwen/Qwen3.5-4B} and \url{https://huggingface.co/Qwen/Qwen3.5-9B}.}
Evaluation uses the original images supplied by each benchmark. All compared methods share the sampling parameters, benchmark prompts, and scoring procedure.

\paragraph{Scoring and comparison settings.}
Tab.~\ref{tab:evaluation_protocol} lists the evaluation tracks, sample counts, and metrics. Following the released scorer, MMVP uses per-question accuracy, CV-Bench averages 2D and 3D accuracy, and POPE uses accuracy. Tab.~\ref{tab:comparison_conditions} and Tab.~\ref{tab:experimental_setup} summarize the supervision conditions and training configuration for both backbones. The OPD and VA-OPD baselines use the Qwen3.5-27B teacher model for both student backbones.
\begin{table*}[tp]
\centering\small
\setlength{\tabcolsep}{3pt}
\begin{tabular*}{\textwidth}{@{\extracolsep{\fill}}lrp{0.27\textwidth}p{0.30\textwidth}@{}}
\toprule
Benchmark & Samples & Evaluation input / track & Reported metric \\
\midrule
V* Bench & 191 & Released benchmark manifest & Accuracy (\%) \\
ZoomBench & 845 & Global-input evaluation & Accuracy (\%) \\
HR-Bench 4K / 8K & 800 / 800 & Separate resolution tracks & Accuracy (\%), separately \\
MME-RealWorld EN / CN & 23609 / 5917 & Separate full language tracks & Accuracy (\%), separately \\
\midrule
MMVP & 300 & Released question manifest & Per-question accuracy (\%) \\
CV-Bench & 2638 & Both 2D and 3D partitions & Mean of 2D and 3D accuracies (\%) \\
MMStar & 1500 & Full evaluation set & Accuracy (\%) \\
POPE & 9000 & Released POPE evaluation set & Accuracy (\%) \\
\bottomrule
\end{tabular*}
\caption{\textbf{Evaluation benchmarks and metrics following Vision-OPD~\cite{yuan2026visionopd}.} The upper block contains six fine-grained tracks; the lower block contains four holdout tasks. Fine Avg. is $\sum_{b=1}^{6} n_b a_b / 32162$, where $a_b$ is accuracy (\%) and $n_b$ is the sample count for each upper-block track. All scores are percentages; scoring follows the released evaluation code.}
\label{tab:evaluation_protocol}
\end{table*}

\begin{table*}[tp]
\centering\small
\setlength{\tabcolsep}{5pt}
\renewcommand{\arraystretch}{1.12}
\begin{tabularx}{\textwidth}{@{}>{\raggedright\arraybackslash}p{0.15\textwidth}>{\raggedright\arraybackslash}p{0.18\textwidth}>{\raggedright\arraybackslash}p{0.20\textwidth}>{\raggedright\arraybackslash}X@{}}
\toprule
\textbf{Method} & \textbf{Student / data} & \textbf{Additional supervision} & \textbf{Teacher and training details} \\
\midrule
Base model & Qwen3.5-4B or 9B & None & No post-training \\
\cmidrule(lr){1-4}
SFT on crop responses & Same backbone and regional pool & Verified crop responses & \textbf{Teacher:} fixed initial backbone.\newline
\textbf{Generation:} sampled crop responses; max 1024 tokens.\newline
\textbf{Filtering:} exclude empty or truncated outputs. \\
\cmidrule(lr){1-4}
GRPO / DAPO & Same backbone and regional pool & Reference answer via verifier & \textbf{Reward:} binary answer correctness.\newline
\textbf{Sampling:} 8 rollouts; temperature \samplingtemperature{}.\newline
\textbf{GRPO:} symmetric clip $\epsilon_{\mathrm{P}}=0.20$.\newline
\textbf{DAPO:} asymmetric lower/upper clip $(0.20,0.28)$; dynamic sampling and length penalty. \\
\cmidrule(lr){1-4}
OPSD & Same backbone and regional pool & Privileged answer/solution context & \textbf{Teacher:} current model with ground-truth answer hint.\newline
\textbf{Objective:} full-vocabulary forward KL; \mbox{pointwise clip 0.05}. \\
\cmidrule(lr){1-4}
Vision-OPD & Same backbone and regional pool & Evidence crop & \textbf{Teacher:} EMA self-teacher; update rate 0.05.\newline
\textbf{Objective:} symmetric crop-conditioned JSD; top-100 tokens plus tail. \\
\midrule
\textbf{\methodname{}} & Same backbone and regional pool & Evidence crop + answer verifier & \textbf{Teacher:} EMA self-teacher; update rate 0.05.\newline
\textbf{Policy:} symmetric clip $\epsilon_{\mathrm{P}}=0.20$.\newline
\textbf{Token weights:} clipped and sequence-centered.\newline
\textbf{Fallback:} verified all-fail OPD with sampled-token JSD. \\
\bottomrule
\end{tabularx}
\caption{\textbf{Conditions for the component training methods.} For each backbone, all trained methods use the same regional question pool for one epoch, with checkpoints selected by performance. Response generation uses non-thinking mode with temperature \samplingtemperature{}, top-$p=\samplingtopp$, and top-$k=\samplingtopk$. Columns summarize the additional supervision and method-specific teacher, generation, filtering, reward, and objective settings. Policy-ratio clipping is symmetric except in DAPO. The base model receives no post-training.}
\label{tab:comparison_conditions}
\end{table*}

\begin{table*}[tp]
\centering
\small
\setlength{\tabcolsep}{3pt}
\renewcommand{\arraystretch}{1.15}
\begin{tabular*}{\textwidth}{@{\extracolsep{\fill}}>{\raggedright\arraybackslash}p{0.30\textwidth}>{\raggedright\arraybackslash}p{0.17\textwidth}>{\raggedright\arraybackslash}p{0.30\textwidth}>{\raggedright\arraybackslash}p{0.17\textwidth}@{}}
\toprule
\textbf{Setting} & \textbf{Configuration} & \textbf{Setting} & \textbf{Configuration} \\
\midrule
Backbones & Qwen3.5-4B / 9B & Optimizer / learning rate & AdamW / $5\times10^{-7}$ \\
\addlinespace[4pt]
Train / evaluation mode & non-thinking / non-thinking & Batch size / updates & 16 questions (128 responses) / 390 updates \\
\addlinespace[4pt]
Train examples / epochs & 6241 / 1 & EMA student update rate $\alpha$ & 0.05 \\
\addlinespace[4pt]
Evaluation datasets / splits & Tab.~\ref{tab:evaluation_protocol} & Weight / policy clipping $\epsilon_w,\epsilon_{\mathrm{P}}$ & $0.20 / 0.20$ (symmetric) \\
\addlinespace[4pt]
Full / crop image-token budget & Dynamic resolution; backbone processor defaults & Fallback / reference strength $\lambda_{\mathrm{F}},\beta$ & $0.10 / 0.02$ \\
\addlinespace[4pt]
Group size $G$ / maximum response length & $8$ / $1024$ tokens & Fallback objective & Sampled-token JSD \\
\addlinespace[4pt]
Sampling temperature / top-$p$ / top-$k$ & $\samplingtemperature\,/\,\samplingtopp\,/\,\samplingtopk$ & GPU type / number & NVIDIA L40S / 8 \\
\addlinespace[4pt]
Fallback activation & Verified fallback & Checkpoint selection & Best performance within one epoch \\
\addlinespace[4pt]
Token-weight centering & Per response; mean one & Advantage variance denominator & $G$ \\
\bottomrule
\end{tabular*}
\caption{\textbf{Experimental configuration for \methodname{}.} The listed settings apply to Qwen3.5-4B and Qwen3.5-9B. Each batch contains 16 questions with eight student responses each (128 responses); training spans one epoch with 390 updates. Sequence centering follows token-weight clipping and preserves both mean one per response and the weight bounds. Verified fallback uses sampled-token JSD after a crop-teacher answer passes verification. Each EMA update mixes 95\% of the previous teacher with 5\% of the current student. Image-token budgets follow the backbone processor defaults with dynamic resolution. Training and evaluation both use non-thinking mode with the listed sampling parameters.}
\label{tab:experimental_setup}
\end{table*}

\section{Details of the 4B Motivation Audit}
\label{sec:motivation_audit_details}
\begin{table*}[tp]
\centering\small
\setlength{\tabcolsep}{4pt}
\begin{tabular*}{\textwidth}{@{\extracolsep{\fill}}lrr@{}}
\toprule
Observed event & Count / denominator & Rate (\%) \\
\midrule
GRPO all-fail: $k_G=0$ & 353 / 1595 & 22.13 \\
GRPO infrequent or absent success: $k_G\leq2$ & 681 / 1595 & 42.70 \\
GRPO uniform rewards: $k_G\in\{0,8\}$ & 582 / 1595 & 36.49 \\
GRPO all-fail, OPD positive: $k_G=0,\ k_O>0$ & 289 / 1595 & 18.12 \\
OPD positive among GRPO all-fail: $k_O>0\mid k_G=0$ & 289 / 353 & 81.87 \\
OPD low-positive among these cases: $1\leq k_O\leq2\mid k_G=0,k_O>0$ & 95 / 289 & 32.87 \\
OPD low-positive in the common cohort: $1\leq k_O\leq2$ & 297 / 1595 & 18.62 \\
GRPO positive, OPD low-success: $k_G\geq1,\ k_O\leq2$ & 292 / 1595 & 18.31 \\
Extreme reverse disagreement: $k_G=8,\ k_O=0$ & 9 / 1595 & 0.56 \\
\bottomrule
\end{tabular*}
\caption{\textbf{Observed limitations of outcome-only and distillation-only training signals.} We compare 1595 randomly selected training samples with uniquely matched records from both methods. $k_G$ and $k_O$ count correct responses out of eight full-image student rollouts in the GRPO and Vision-OPD runs. Rows report overlapping events with their respective denominators.}
\label{tab:motivation_audit}
\end{table*}

\begin{table*}[tp]
\centering\small
\setlength{\tabcolsep}{4pt}
\begin{tabular*}{\textwidth}{@{\extracolsep{\fill}}lrccc@{}}
\toprule
Cohort & $N$ & $k_O>0\mid k_G=0$ & $1\leq k_O\leq2\mid k_G=0,k_O>0$ & $k_O\leq2\mid k_G\geq1$ \\
\midrule
All common questions & 1595 & 289/353 (81.87\%) & 95/289 (32.87\%) & 292/1242 (23.51\%) \\
All 16 parsed; no known truncation & 1102 & 195/241 (80.91\%) & 64/195 (32.82\%) & 183/861 (21.25\%) \\
Four distinct option texts & 1563 & 286/350 (81.71\%) & 92/286 (32.17\%) & 287/1213 (23.66\%) \\
\bottomrule
\end{tabular*}
\caption{\textbf{Sensitivity of the observational audit.} $N$ is the number of matched questions in each cohort; $k_G$ and $k_O$ count correct GRPO and Vision-OPD responses out of eight, as in Tab.~\ref{tab:motivation_audit}. Each rate cell reports the event count divided by the conditioning-set size, followed by the percentage. The final column conditions on $k_G\geq1$, giving $292/1242=23.51\%$ for all common questions; the corresponding rate over all 1595 questions is 18.31\%. ``No known truncation'' excludes recorded GRPO truncations; truncation flags are unavailable for Vision-OPD. The subsets overlap.}
\label{tab:motivation_audit_sensitivity}
\end{table*}

The audit compares 1595 randomly selected training samples using saved GRPO and Vision-OPD Qwen3.5-4B responses. Each sample has eight full-image responses from each method. Both runs initialize from Qwen3.5-4B and sample at temperature \samplingtemperature{}, top-$p=\samplingtopp$, and top-$k=\samplingtopk$, with thinking disabled. GRPO uses learning rate $5\times10^{-7}$, data seed 42, uniform token advantages, a fixed token normalizer of 128, and reference KL coefficient 0.02. Vision-OPD uses learning rate $2\times10^{-6}$, a different data order, an EMA crop teacher, and no reference KL. The audit uses eight responses per question; benchmark evaluation uses one.

\paragraph{Matching and scoring.}
The source dataset contains 6241 rows. We match the sampled questions by exact prompt and reference answer, retaining unique source mappings. Each question in the comparison has one response group per method and corresponds to a distinct original image. The same repository MCQ verifier scores both runs against the dataset's reference answers. GRPO re-scoring agrees with the stored rewards; Vision-OPD responses are scored retrospectively.

\paragraph{Sensitivity to response quality and option validity.}
Tab.~\ref{tab:motivation_audit_sensitivity} reports two subsets of the 1595-sample comparison. The fully parsed/no-known-truncation subset contains 1102 questions, and the subset with four distinct option texts contains 1563. The first subset requires all 16 responses to be parsed and excludes recorded GRPO truncations; Vision-OPD does not retain truncation flags. Both subsets preserve the main patterns: Vision-OPD produces correct responses on roughly four-fifths of GRPO all-fail questions, while about one-third of these groups still contain only one or two successes.

\section{Additional Benchmark Results and Case Studies}
\label{sec:benchmark_case_sources}
\label{sec:additional_results}
\begin{figure*}[tp]
\centering
\setlength{\abovecaptionskip}{6pt}
\includegraphics[width=\textwidth,trim=0 0 0 6bp,clip]{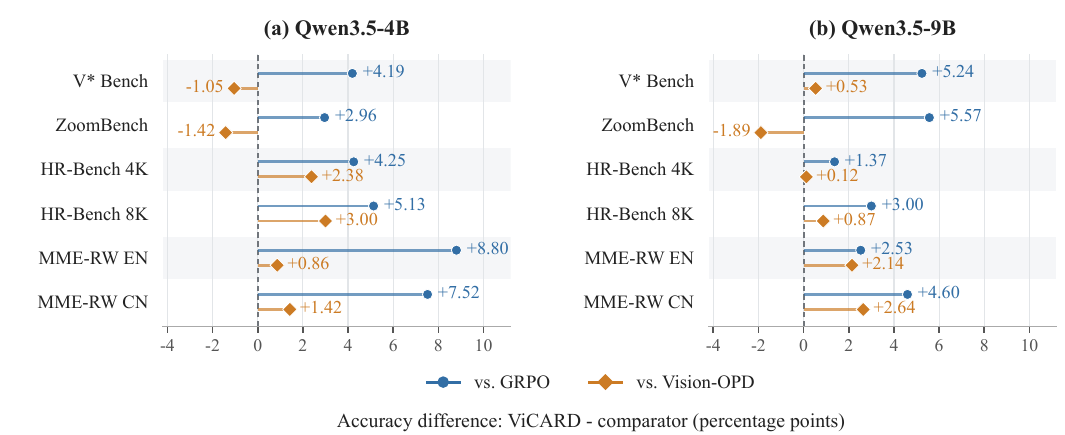}
\caption{\textbf{Benchmark-wise performance differences.} Accuracy differences between \methodname{} and GRPO (blue circles) or Vision-OPD (orange diamonds) on the six fine-grained evaluation tracks for (a) Qwen3.5-4B and (b) Qwen3.5-9B. Values are computed from Tab.~\ref{tab:main_benchmarks} and labeled in percentage points; positive values favor \methodname{}. Both panels use the same axis range, and the dashed vertical line marks equal accuracy. MME-RW denotes MME-RealWorld.}
\label{fig:main_performance_gains}
\end{figure*}

\begin{figure*}[tp]
\centering
\setlength{\abovecaptionskip}{6pt}
\includegraphics[width=0.78\textwidth,trim=0 8bp 0 3bp,clip]{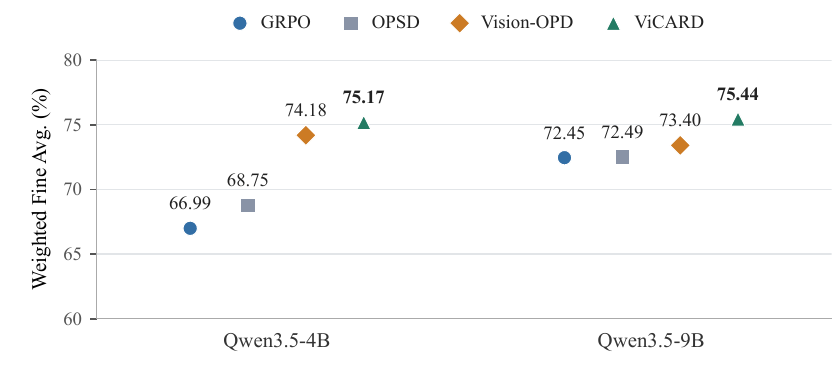}
\caption{\textbf{Fine-grained performance across the two student backbones.} Sample-count-weighted accuracy over the six fine-grained evaluation tracks in Tab.~\ref{tab:main_benchmarks}. Points compare GRPO, OPSD, Vision-OPD, and \methodname{} under the same regional training setting and evaluation protocol, with each score labeled above its marker. The vertical axis spans 60--80\%. \methodname{} achieves 75.17\% at 4B and 75.44\% at 9B.}
\label{fig:main_backbone_comparison}
\end{figure*}

\begin{figure*}[tp]
\centering
\includegraphics[width=\textwidth]{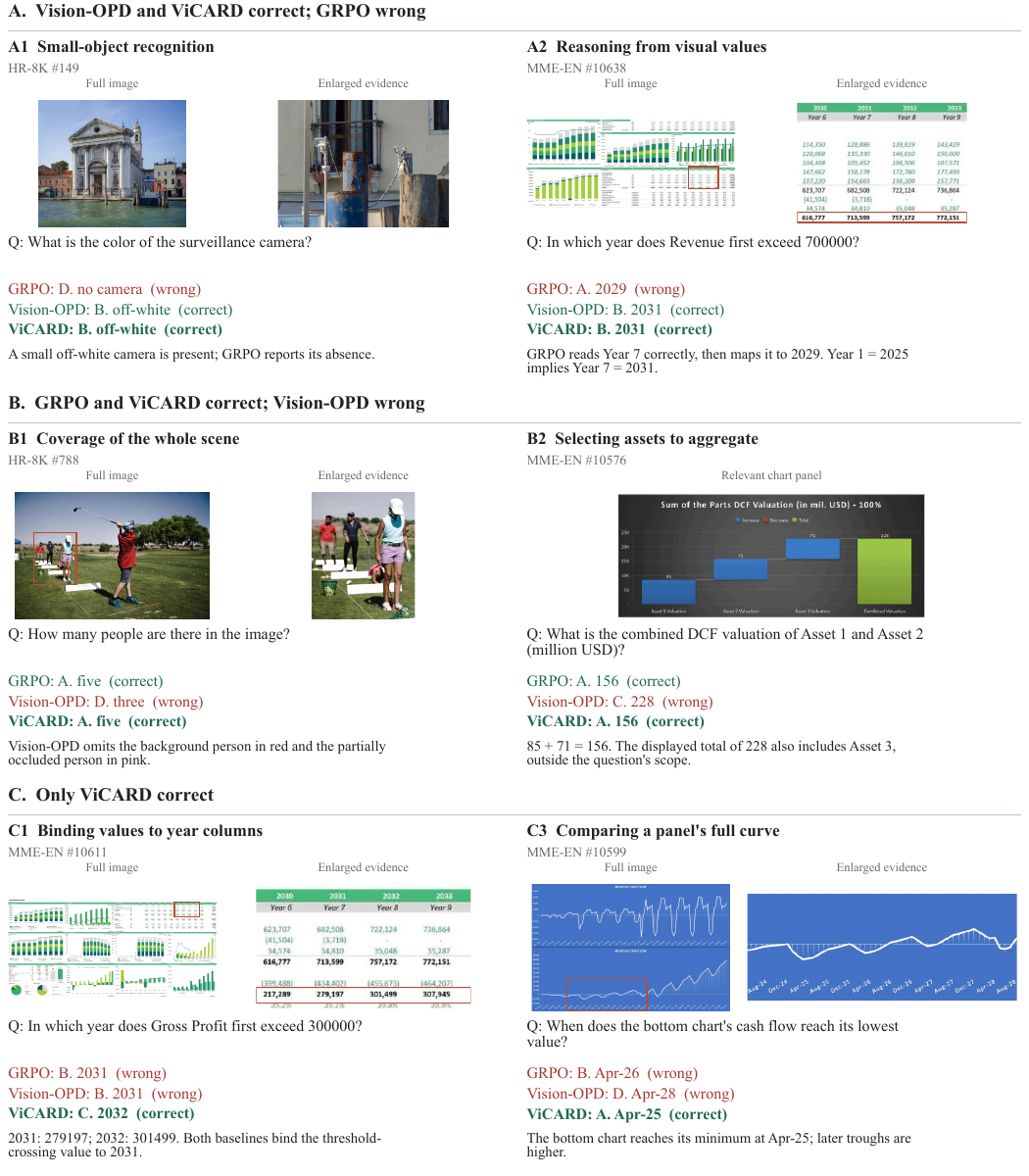}
\caption{\textbf{Complementary visual and reasoning capabilities in six 4B benchmark cases.} Group A pairs \methodname{} with Vision-OPD, group B pairs it with GRPO, and group C contains cases answered correctly only by \methodname{}. HR-8K and MME-EN denote HR-Bench 8K and MME-RealWorld EN; numbers identify benchmark questions. Questions are shortened for display, and answers summarize the saved responses. Boxes and enlarged views highlight evidence for presentation; evaluation uses the original full images.}
\label{fig:additional_showcases}
\end{figure*}

\subsection{Benchmark and Backbone Trends}
Fig.~\ref{fig:main_performance_gains} and Fig.~\ref{fig:main_backbone_comparison} complement Sec.~\ref{sec:vrlsd_performance} by showing where advantages concentrate and how the training strategies behave across the two backbone sizes.

\begin{samepage}
\paragraph{Benchmark profiles.}
In Fig.~\ref{fig:main_performance_gains}, \methodname{}'s gains over GRPO peak on MME-RealWorld at 4B and shift toward extreme small-object benchmarks (ZoomBench and $V^*$ Bench) at 9B, suggesting that once the larger backbone acquires stronger baseline reasoning, visual credit assignment primarily unlocks ultra-fine perceptual recovery. Against Vision-OPD, \methodname{}'s advantage consistently widens as resolution increases (HR-Bench 8K vs.\ 4K) and on complex real-world queries (MME-RealWorld CN/EN), where global context and multi-step reasoning are required alongside local detail recognition.\par
\end{samepage}

\paragraph{Backbone trends and weighting.}
Fig.~\ref{fig:main_backbone_comparison} shows larger gains from 4B to 9B for GRPO and OPSD, a slight decline for Vision-OPD, and nearly unchanged weighted accuracy for \methodname{}. The 4B \methodname{} model already exceeds all three 9B comparators plotted in the figure. Within \methodname{}, the larger backbone improves on five of the six tracks, while its MME-RealWorld EN score declines slightly (Tab.~\ref{tab:main_benchmarks}). Because this track carries most of the Fine Avg. weight (Tab.~\ref{tab:evaluation_protocol}), its decline offsets part of the gains on the other tracks, explaining the small aggregate change between backbones.

\subsection{Complete Case Analyses}
\paragraph{Case sources and presentation.}
Fig.~\ref{fig:additional_showcases} uses saved benchmark responses from the selected Qwen3.5-4B GRPO, Vision-OPD, and \methodname{} checkpoints. The selected HR-Bench 8K question IDs are 149 and 788; the MME-RealWorld EN IDs are 10638, 10576, 10611, and 10599. Each case contains one saved response per method. Case selection uses the final answer option, checked against the original image and the full response. Fig.~\ref{fig:additional_showcases} presents the final answers and concise explanations; the magnified views and boxes are added for display.

\paragraph{From visual evidence to a correct answer.}
Fig.~\ref{fig:additional_showcases} examines six Qwen3.5-4B benchmark cases, grouped by which methods answer correctly. The first pair (A1--A2) is solved by Vision-OPD and \methodname{} but missed by GRPO. In A1, both methods identify an off-white surveillance camera in a high-resolution waterfront scene, while GRPO concludes that no camera is present. A2 isolates a different failure: GRPO correctly reads the revenue values and identifies Year~7 as the first year above 700000, yet maps that year to 2029 despite stating that Year~1 is 2025. \methodname{} correctly associates the value 713599 with 2031, as does Vision-OPD. The required conversion is $2025+(7-1)=2031$. Together, these cases show improvements in both identifying fine visual evidence and reasoning from correctly extracted values.

\paragraph{Preserving scene coverage and question context.}
The second pair (B1--B2) is solved by GRPO and \methodname{} but missed by Vision-OPD. In B1, Vision-OPD counts three people, omitting a person in red in the background and a partially occluded person in pink. GRPO and \methodname{} explicitly account for all five people. This requires coverage of the full scene as well as recognition of a small, occluded subject. In B2, Vision-OPD correctly reads the displayed total valuation of 228 million USD, but the question asks for Asset~1 and Asset~2 only. GRPO and \methodname{} select the two requested values and compute $85+71=156$, excluding Asset~3. \methodname{} therefore combines local visual recognition with the surrounding scene and the scope specified by the question, supporting both complete evidence collection and appropriate aggregation.

\paragraph{Joint evidence binding and reasoning.}
Only \methodname{} answers the final pair (C1 and C3) correctly. In C1, both baselines identify 301499 as the gross-profit value crossing 300000 but assign it to 2031. GRPO's explanation shifts the first populated value into the empty START/Year~0 column. \methodname{} aligns 279197 with 2031 and 301499 with 2032, then selects 2032. In C3, the question specifies the bottom cash-flow chart. GRPO selects Apr-26 and Vision-OPD selects Apr-28, whereas \methodname{} identifies the bottom chart's global minimum at Apr-25. These cases require preserving relationships between values and headers, or between a curve, its axis, and the requested panel. Together with the preceding pairs, they illustrate how \methodname{} unites fine-detail recognition, broader visual context, and reasoning over the selected evidence. The token-level diagnostics in Appendix~\ref{sec:token_diagnostic_protocol} complement these answer-level case studies.

\section{Token-Importance Diagnostics}
\label{sec:token_diagnostic_protocol}
These diagnostics expand the token-level observations in Sec.~\ref{sec:vrlsd_motivation} and examine how crop-to-full token log-ratios change after training.
\begin{figure*}[tp]
\centering
\begin{subfigure}[t]{\textwidth}
\centering
\includegraphics[width=0.98\linewidth]{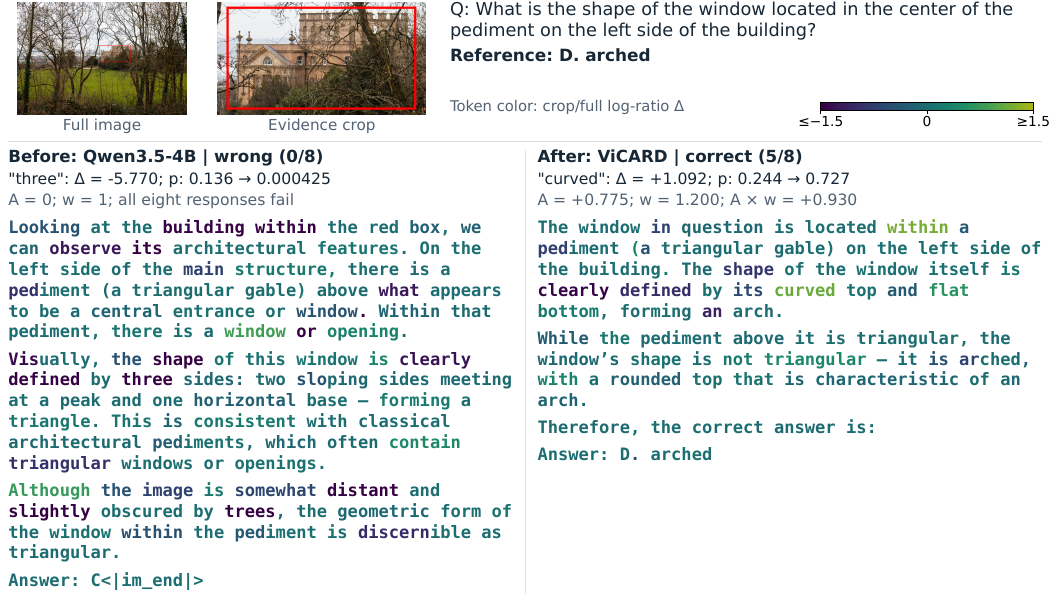}
\caption{Window shape: triangular $\rightarrow$ arched.}
\label{fig:token_importance_window}
\end{subfigure}
\par\smallskip
\begin{subfigure}[t]{\textwidth}
\centering
\includegraphics[width=0.98\linewidth]{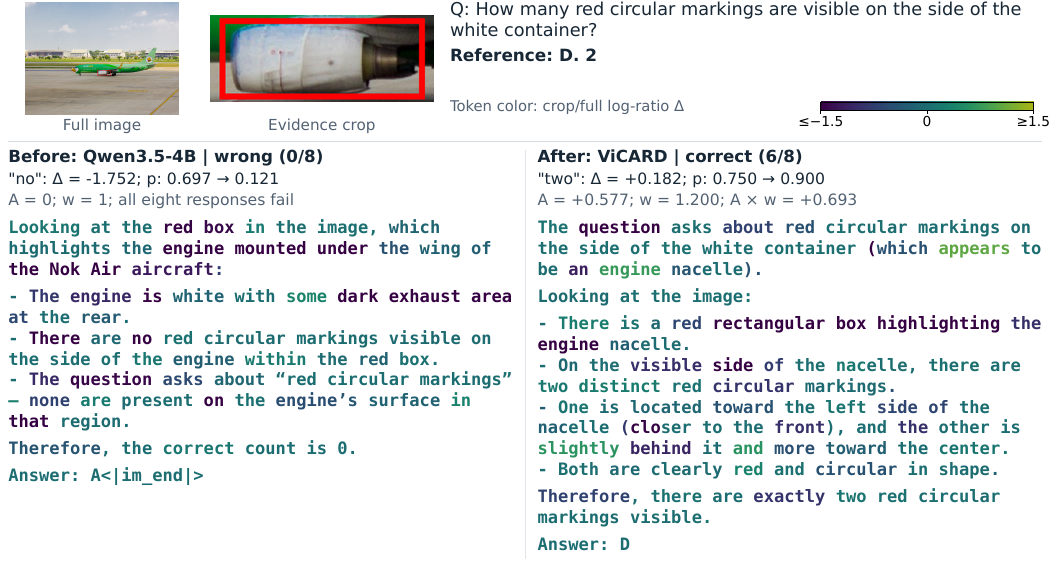}
\caption{Red circular markings: none $\rightarrow$ two.}
\label{fig:token_importance_markings}
\end{subfigure}
\captionsetup{labelformat=tokenimportanceab}
\caption{\textbf{Wrong-to-correct visual reasoning on the same questions.} Complete base Qwen3.5-4B (left) and \methodname{} (right) responses are colored by $\Delta=\log p_{\mathrm{crop}}-\log p_{\mathrm{full}}$, saturating at $\pm1.5$. Counts give correct answers out of eight responses; annotations give token probabilities and sequence-centered weights. Each model scores its own generated prefixes. Panels (a,b) show window shape and red markings; continued panels (c--e) revisit three cases from Fig.~\ref{fig:motivation_cases}. Scoring and response-selection details are in Sec.~\ref{sec:token_importance_protocol}.}
\label{fig:token_importance}
\end{figure*}

\begin{figure*}[tp]
\ContinuedFloat
\setcounter{subfigure}{2}
\centering
\begin{subfigure}[t]{\textwidth}
\centering
\includegraphics[width=0.98\linewidth]{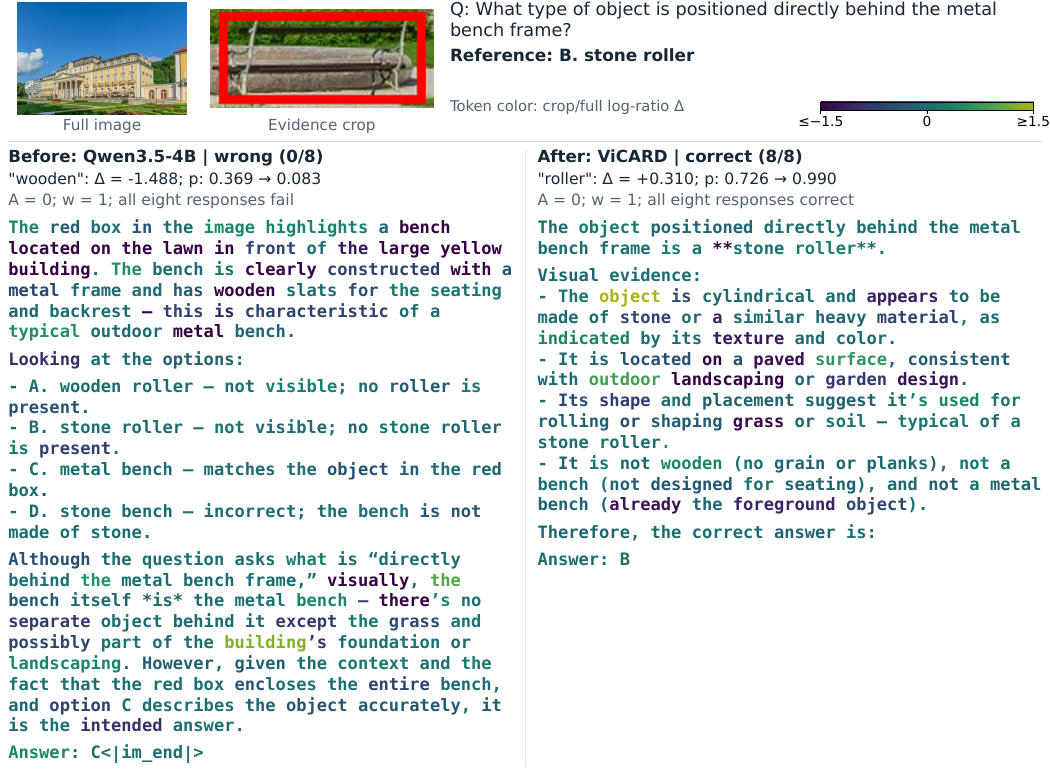}
\caption{Stone roller (ID 2669): a bench interpretation $\rightarrow$ stone roller.}
\label{fig:token_importance_roller}
\end{subfigure}
\captionsetup{labelformat=tokenimportancec}
\caption{\textbf{Wrong-to-correct visual reasoning (continued).} Panel (c) reuses the image and question in Fig.~\ref{fig:motivation_cases}\subref{fig:motivation_case_object}, with newly generated base/\methodname{} responses. The crop lowers the probability of ``wooden'' in the base response and raises that of ``roller'' in the trained response. Both selected groups have uniform rewards, so $A=0$ and $w=1$; text colors show the raw token log-ratios under the common scale.}
\label{fig:token_importance_roller_page}
\end{figure*}

\begin{figure*}[tp]
\ContinuedFloat
\setcounter{subfigure}{3}
\centering
\begin{subfigure}[t]{\textwidth}
\centering
\includegraphics[width=0.98\linewidth]{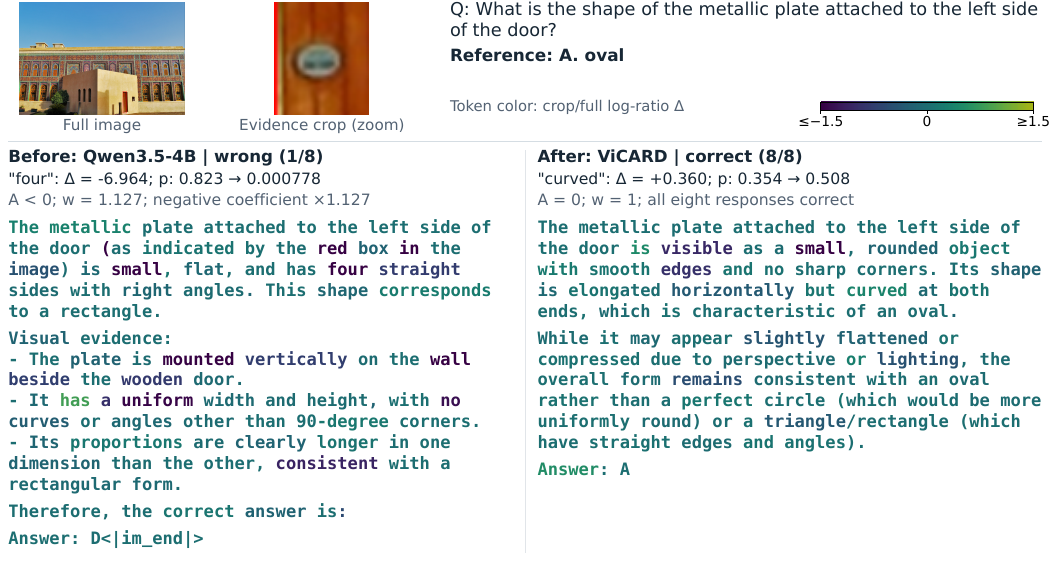}
\caption{Oval plate (ID 3092): four straight sides $\rightarrow$ a curved contour.}
\label{fig:token_importance_plate}
\end{subfigure}
\par\smallskip
\begin{subfigure}[t]{\textwidth}
\centering
\includegraphics[width=0.98\linewidth]{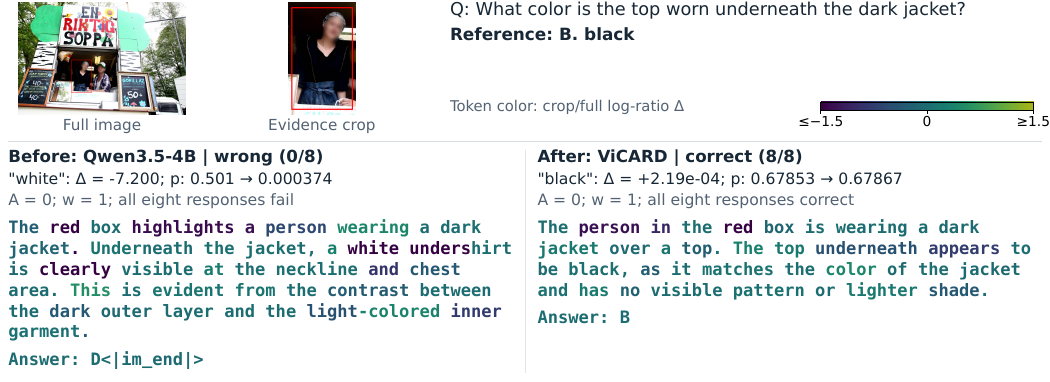}
\caption{Clothing color (ID 3035): white $\rightarrow$ black.}
\label{fig:token_importance_clothing}
\end{subfigure}
\captionsetup{labelformat=tokenimportancede}
\caption{\textbf{Wrong-to-correct visual reasoning (continued).} Panels (d,e) reuse the cases in Fig.~\ref{fig:motivation_cases}(b,d). In the plate's base group, one of eight responses is correct; the selected incorrect answer has $A<0$, and the centered weight $w=1.127$ increases the magnitude of the negative coefficient on ``four''. The trained groups in panels (c--e) are 8/8 correct, giving $A=0$ and $w=1$. The plate crop is enlarged for display; scoring uses the supplied crop.}
\label{fig:token_importance_plate_clothing_page}
\end{figure*}

\begin{figure*}[!tp]
\centering
\includegraphics[width=\textwidth,trim=0 10bp 0 10bp,clip]{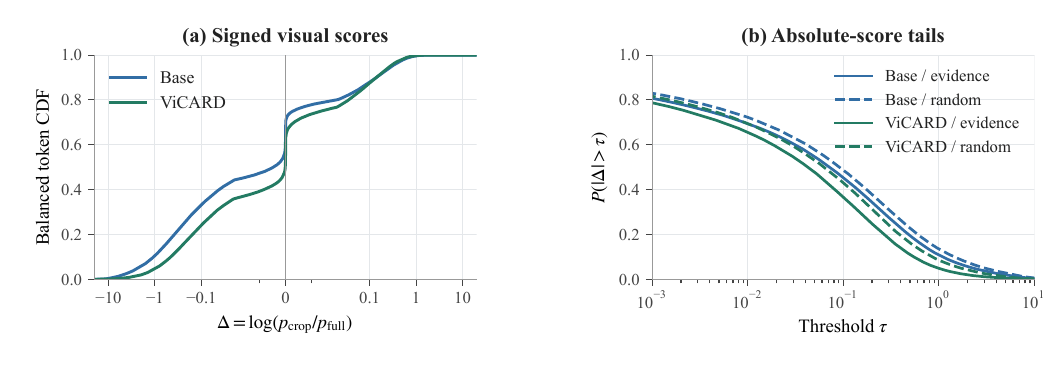}
\caption{\textbf{Token log-ratio distributions before and after training.} Base Qwen3.5-4B and \methodname{} independently generate eight responses for each of the same 200 training questions. We score reasoning tokens preceding the final answer, with equal mass per image and response. (a) CDF of the crop/full log-ratio $\Delta$ on a symmetric log scale. (b) Tail probabilities of $|\Delta|$ using evidence crops (solid) or area-matched random crops (dashed). These log-ratios are used to construct token weights.}
\label{fig:token_importance_distribution}
\end{figure*}

\begin{figure*}[!tp]
\centering
\includegraphics[width=\textwidth,trim=0 14bp 0 10bp,clip]{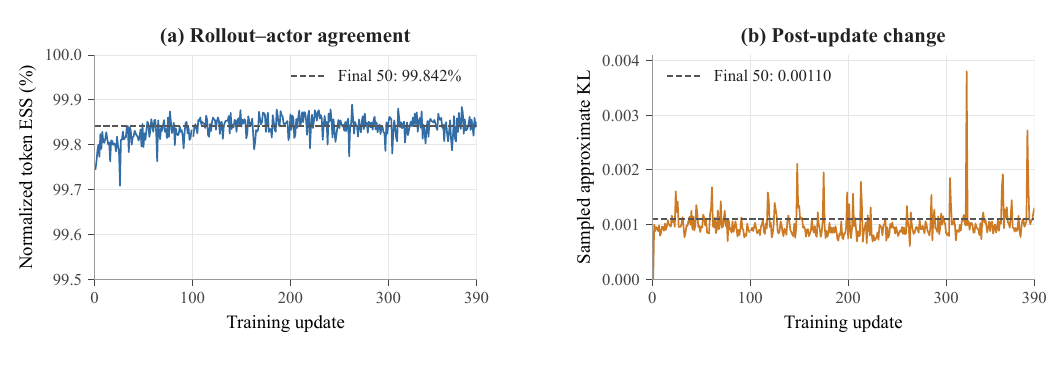}
\caption{\textbf{On-policy diagnostics during 4B \methodname{} training.} (a) Normalized token effective sample size, computed from pre-update actor/rollout probability ratios clamped to $[0.5,2]$. (b) Sampled approximate KL after rescoring the same responses following each optimizer update. Dashed lines mark arithmetic means over updates 341--390. The curves cover 389 recorded updates and are disconnected at the missing update 100.}
\label{fig:on_policy_diagnostics}
\end{figure*}

\subsection{Paired Token-Level Examples}
\label{sec:token_importance_cases}
\paragraph{Visual evidence provides nonuniform token-level information.}
The left columns of Fig.~\ref{fig:token_importance}(a,b) show two visual reasoning errors from base Qwen3.5-4B. In the window example, the response attributes the surrounding pediment's triangular shape to the window and describes it as having ``three sides''. Scoring the same token and prefix using the full image as input versus the evidence crop reduces the probability of ``three'' from $0.136$ to $4.25\times10^{-4}$ ($\Delta=-5.770$). The text colors encode this crop-to-full log-ratio, connecting each visual claim to the change in its token support. In the markings example, the response denies the presence of red circular markings; the crop reduces the probability of the first ``no'' from $0.697$ to $0.121$ ($\Delta=-1.752$). Both questions have zero correct answers in eight base-model responses, so their group-relative advantages vanish.

\paragraph{From incorrect visual judgments to evidence-based answers.}
Fig.~\ref{fig:token_importance}(a,b) pairs complete responses to the same two training questions before and after \methodname{} training. In panel~\subref{fig:token_importance_window}, the trained model separates the triangular pediment from the window's curved top and correctly answers ``arched''. Correct responses increase from 0/8 to 5/8 in the saved groups. The evidence crop raises the probability of ``curved'' from $0.244$ to $0.727$ ($\Delta=1.092$); its centered weight is $1.200$, increasing the positive coefficient from $A_i=0.775$ to $0.930$. In panel~\subref{fig:token_importance_markings}, the model replaces the denial of red markings with an explicit count of two, and correct responses increase from 0/8 to 6/8. The probability of ``two'' rises from $0.750$ to $0.900$ under the crop, yielding $w=1.200$ and increasing its coefficient from $0.577$ to $0.693$. These pairs connect improved visual judgments with token credit that emphasizes the relevant geometric and counting cues. Sequence centering redistributes this credit while preserving each response's mean outcome coefficient.

\paragraph{Question-relevant evidence drives the token contrast.}
Random-crop scores further localize the source of the contrast. For the base model's erroneous ``no'', the probability is $0.697$ under the full image, $0.121$ under the evidence crop, and $0.709$ under the random crop. For the trained model's correct ``curved'', the corresponding probabilities are $0.244$, $0.727$, and $0.203$. The relevant crop suppresses the incorrect denial and supports the correct window geometry, while the random crop provides neither change. Scoring and response-selection details are given in Sec.~\ref{sec:token_importance_protocol}.

\paragraph{Object identity and shape in the motivation cases.}
Panels~\subref{fig:token_importance_roller} and~\subref{fig:token_importance_plate} revisit the stone-roller and oval-plate questions in Fig.~\ref{fig:motivation_cases}, using newly generated base and trained-model responses. Their correct-answer counts rise from 0/8 to 8/8 and from 1/8 to 8/8, respectively. For the stone roller, the crop lowers support for ``wooden'' from $0.369$ to $0.083$ ($\Delta=-1.488$), while increasing support for ``roller'' in the trained response from $0.726$ to $0.990$ ($\Delta=0.310$). For the oval plate, the base response's ``four'' has full-image probability $0.823$, but crop probability $7.78\times10^{-4}$ ($\Delta=-6.964$); the random crop retains probability $0.866$. Its centered weight is $1.127$, increasing the magnitude of its negative outcome coefficient by 12.7\%. The trained response instead describes the plate as ``curved'', with crop support increasing from $0.354$ to $0.508$.

\paragraph{Color correction with agreement between views.}
Panel~\subref{fig:token_importance_clothing} revisits the clothing-color question in Fig.~\ref{fig:motivation_cases}\subref{fig:motivation_case_clothing}. The base response's ``white'' falls from probability $0.501$ under the full image to $3.74\times10^{-4}$ under the crop ($\Delta=-7.200$). After training, the correct ``black'' has nearly equal full-image and crop probabilities, $0.67853$ and $0.67867$, whereas its random-crop probability is $0.00362$. Correct responses rise from 0/8 to 8/8. Here the near-zero log-ratio $\Delta$ reflects agreement on the correct color. The trained groups in panels (c--e) are all-success, giving $A=0$ and $w=1$; the colors show crop-to-full differences in token support.

\subsection{Scoring and Rendering Protocol}
\label{sec:token_importance_protocol}
\paragraph{Paired responses and visual conditions.}
Panels (a,b) use an export covering 200 regional training images, with eight responses per question from base Qwen3.5-4B and \methodname{} at step~390. They show questions 184 (window shape) and 2521 (red circular markings), using trial~1 from each checkpoint. Panels (c--e) use a separate four-question export of the images in Fig.~\ref{fig:motivation_cases}; the displayed wrong-to-correct pairs are questions 2669, 3092, and 3035, using trials 2, 3, and 2, respectively. These diagnostic groups compare the initial and trained checkpoints; Fig.~\ref{fig:motivation_cases} reports the GRPO/Vision-OPD audit groups for the same images. Each pair shares its image, question, options, and reference answer, with independently generated responses before and after training. Generation uses temperature $\samplingtemperature$, top-$p=\samplingtopp$, and top-$k=\samplingtopk$, with at most 512 new tokens for (a,b) and 1024 for (c--e). Each frozen checkpoint scores its own exact token IDs and prefixes under the marked full image, evidence crop, and random crop at scoring temperature $1.0$. The random-crop overlap with the target box is $6.92\times10^{-4}$ IoU for the window and zero for the markings. The plate image includes a display enlargement of its supplied evidence crop.

\paragraph{Answer readout and group outcomes.}
For the final-answer audit, we remove trailing special-token strings from an audit copy, read the last anchored \texttt{Answer: A--D} line, and compare it with the reference option. For panels (a--e), the base groups contain 0, 0, 0, 1, and 0 correct answers; the trained groups contain 5, 6, 8, 8, and 8, respectively. All selected base responses are incorrect and all selected trained responses are correct. The trained groups in panels (a,b) have mixed rewards and positive advantages for the selected answers; those in panels (c--e) have uniform rewards and zero advantages. Among the base groups in panels (c--e), only the plate has mixed rewards.

\paragraph{Scores, centering, and rendering.}
Both columns of every pair in Fig.~\ref{fig:token_importance} color tokens by the within-checkpoint log-ratio $\Delta_{i,t}=\log p_{\mathrm{crop}}-\log p_{\mathrm{full}}$. The common color scale saturates at $[-1.5,1.5]$, and numerical annotations retain the full scores. We compute group advantages from the saved binary rewards using the population variance in Eq.~\eqref{eq:vrlsd_group_advantage}, with denominator $G$ and $\varepsilon_{\mathrm{num}}=10^{-6}$. Weight annotations use the exported sequence-centered weights, verified against Eq.~\eqref{eq:vrlsd_credit} with $\epsilon_w=0.2$, and report their products with these advantages where shown. The plate panel reports the relative multiplier on its negative coefficient. Centering covers all valid response tokens and maintains mean one and bounds $[0.8,1.2]$.

Each displayed answer is complete, including its final answer and any decoded end marker. Exact character offsets map visible characters to the original token scores; line wrapping preserves paragraph and list boundaries. Whitespace-only and zero-width tokens participate in weight computation but have no colored glyphs.

\subsection{Token Log-Ratio Distributions after Training}
\label{sec:token_distribution_changes}
Using the original 200-image export, we compare 1600 independently generated responses per checkpoint. We analyze $\Delta$ on reasoning tokens preceding the final \texttt{Answer:} section, excluding special-token text. To prevent longer trajectories from dominating the statistics, we weight tokens uniformly within each response, responses uniformly within each question, and questions uniformly across the 200-image cohort. Fig.~\ref{fig:token_importance_distribution} shows that the distribution concentrates closer to zero after training. Mean $|\Delta|$ decreases from $0.478$ to $0.240$ (49.8\%), its 95th percentile falls from $2.293$ to $0.985$, and $P(|\Delta|>1)$ drops from 10.84\% to 4.91\%. The paired mean change is $-0.238$, with a 95\% image-bootstrap interval of $[-0.261,-0.216]$ from 10000 resamples. Mean absolute log-ratios under random crops also decrease, from $0.596$ to $0.372$. These shifts indicate closer full-image/crop agreement on the trained model's own generated reasoning.

\section{On-Policy Training Diagnostics}
\label{sec:on_policy_diagnostics}
To complement the token log-ratio analysis, we examine the 4B training run by comparing rollout and pre-update actor probabilities on sampled tokens, then rescoring the same full-image responses after each optimizer update (Fig.~\ref{fig:on_policy_diagnostics}). Over the final 50 updates, the step-averaged actor/rollout probability ratio is $1.00005$, and the normalized token effective sample size is 99.842\%. The latter uses ratios clamped to $[0.5,2]$; only 0.0051\% of tokens fall outside this interval on average. With $d=\log p_{\mathrm{after}}-\log p_{\mathrm{before}}$, the sampled approximate-KL diagnostic averages $e^d-1-d$ over response tokens and has a step-averaged value of $0.00110$. The corresponding mean log-probability-change RMS is $0.0465$. Together, these diagnostics support near-on-policy optimization through the end of training.

\section{Extended Related Work}
\label{sec:extended_related_work}
This section expands the overview in Sec.~\ref{sec:related_work}, detailing the mechanisms most closely related to \methodname{}.

\paragraph{Reinforcement learning for visually grounded reasoning.}
Group relative policy optimization (GRPO)~\cite{shao2024deepseekmath} provides a foundation for learning from verifiable outcome rewards, with subsequent work adapting reinforcement learning to visual reasoning. VL-Rethinker~\cite{wang2025vlrethinker} introduces selective sample replay to address vanishing advantages and explicitly encourages self-reflection, while NoisyRollout~\cite{liu2025noisyrollout} mixes trajectories from clean and distorted images to improve exploration. For text reasoning, BREAD~\cite{zhang2025bread} addresses sparse success through partial expert guidance and branched rollouts, and Wang et al.~\cite{wang2025hightokenentropy} identify high-entropy reasoning tokens as effective targets for policy-gradient updates. A closely related direction incorporates perceptual signals into optimization. PAPO~\cite{wang2025papo} adds an implicit perception loss and entropy regularization; VPPO~\cite{huang2026vppo} reweights trajectory advantages by visual dependency and selects perceptually important tokens for policy updates. PEPO~\cite{li2026pepo} combines a hidden-state perception prior with token entropy to construct token-level advantages. These methods establish visual grounding and token-level credit assignment as important dimensions of multimodal RL. \methodname{} builds on this direction by using crop-to-full prediction differences from a self-teacher to modulate outcome advantages, together with a separate distillation objective for verified all-fail groups.

\paragraph{Fine-grained perception and visual evidence acquisition.}
Methods that ``think with images'' acquire additional evidence during reasoning. V*~\cite{wu2024vstar} guides visual search with a language model, while $\mathrm{DC}^{2}$~\cite{wang2025hrbench} decomposes high-resolution images and retrieves question-relevant patches. DeepEyes~\cite{zheng2026deepeyes} learns active visual perception through RL, Pixel Reasoner~\cite{wang2025pixelreasoner} introduces pixel-space operations with curiosity-driven training, and Thyme~\cite{zhang2025thyme} learns to generate executable code for image manipulation and computation. An alternative is to internalize regional information during training. Zooming without Zooming~\cite{wei2026zwz} uses strong teachers on cropped regions to construct supervision for full-image learning. \methodname{} shares this training-time use of regional evidence: crops guide the learning objective, while inference uses the full-image policy without an additional crop-and-query loop.

\paragraph{On-policy distillation and self-distillation.}
STaR~\cite{zelikman2022star} and ReST$^{\mathrm{EM}}$~\cite{singh2024restem} iteratively fine-tune on successful model-generated solutions, establishing self-training as a route to stronger reasoning. GKD~\cite{agarwal2024opd} and MiniLLM~\cite{gu2024minillm} develop distillation on student-generated trajectories, addressing the mismatch between training prefixes and those encountered at inference. Recent work studies reward extrapolation in generalized OPD~\cite{yang2026exopd}, teacher-entropy-dependent divergence objectives~\cite{jin2026entropyopd}, and exploration and supervision reliability across language and multimodal models~\cite{hou2026uniopd}. Self-distillation obtains teacher supervision from the model itself under a more informative context: OPSD~\cite{zhao2026opsd} conditions on privileged solutions, SDPO~\cite{hubotter2026sdpo} uses feedback or successful attempts, and U-OPSD~\cite{li2026uopsd} constructs pseudo-solutions through self-consistency. SDFT~\cite{shenfeld2026sdft} uses a demonstration-conditioned self-teacher for on-policy learning while preserving previously acquired capabilities. Recent analysis reports degradation of long-trace thinking models under privileged self-distillation~\cite{kaur2026rethinkingopsd}. \methodname{} couples crop-based self-supervision with verified outcome rewards, using the observed group outcomes to determine how the teacher signal is applied.

\paragraph{Visual on-policy distillation.}
Two recent approaches are particularly close to ours. Vision-OPD~\cite{yuan2026visionopd} identifies a regional-to-global perception gap and matches a crop-conditioned self-teacher to a full-image student along student-generated prefixes. Its training objective requires neither answer labels nor a reward verifier. VA-OPD~\cite{liu2026vaopd} measures visual advantage through a rectified teacher log-probability difference between original and detail-degraded images. It uses this signal to reweight rollouts and separately normalize distillation losses over high- and low-advantage token groups, with the original-image teacher supplying the targets. \methodname{} draws on regional self-supervision and view-dependent token scoring, but assigns them distinct roles according to the reward group. For mixed-reward groups, a signed crop-to-full score scales GRPO advantages while preserving their outcome-derived signs. When every full-image response fails, these advantages vanish; we then enable crop-teacher OPD only if a separately generated crop-teacher answer passes verification. Distillation remains on the failed student's prefixes, and the verification response serves only to determine eligibility. Thus, our formulation combines reward-anchored visual credit assignment with a verified fallback through group-dependent supervision.

\end{document}